\documentclass[preprint,12pt]{elsarticle}

\usepackage[utf8]{inputenc}
\usepackage{amsmath,amsfonts,amssymb,amsthm,bm,mathtools}
\usepackage[dvipsnames]{xcolor}
\usepackage{hyperref}
\usepackage{siunitx}
\usepackage{graphicx}
\usepackage{float}
\usepackage{subcaption}
\usepackage{tabularray}
\usepackage{cleveref}
\usepackage{bm}
\newcommand{\mathbm}[1]{\bm{#1}}
\usepackage{subcaption}

\crefname{figure}{Fig.}{Figs.}
\Crefname{figure}{Fig.}{Figs.}
\crefname{table}{Tab.}{Tabs.}
\Crefname{table}{Tab.}{Tabs.}
\crefname{equation}{Eq.}{Eqs.}
\Crefname{equation}{Eq.}{Eqs.}
\crefname{section}{Sec.}{Secs.}
\Crefname{section}{Sec.}{Secs.}

\UseTblrLibrary{booktabs}

\DeclarePairedDelimiterX{\infdivx}[2]{(}{)}{%
  #1\;\delimsize\|\;#2%
}

\begin{document}
\begin{frontmatter}

\title{Uncertainty-aware Multi-fidelity Closure via Conditional Normalizing Flows}
\author[PNNL]{Jice Zeng}
\author[PNNL]{Shady E. Ahmed}
\author[PNNL]{David Barajas-Solano\corref{label1}}
\ead{David.Barajas-Solano@pnnl.gov}
\cortext[label1]{Corresponding author}
\author[PNNL]{Panos Stinis}

\affiliation[PNNL]{organization={Advanced Computing, Mathematics, and Data Division, Pacific Northwest National Laboratory},
city={Richland},
postcode={99352},
state={WA, USA}}

\begin{abstract}
Reduced-order models (ROMs) provide efficient surrogates for complex multiscale systems, but their predictive accuracy is often compromised by truncation errors and the inadequate representation of interactions between resolved and unresolved scales. The missing effect of truncated (unresolved) scales on ROM (resolved) scales is often denoted as the closure problem. In this work, we formulate ROM closure modeling as a multi-fidelity (MF) learning problem and propose an uncertainty-aware MF framework based on conditional normalizing flows to enhance ROM predictive accuracy. The proposed approach learns a probabilistic mapping from low-fidelity (LF) ROM coefficients to high-fidelity (HF) coefficients, thereby improving predictive fidelity while quantifying the uncertainty associated with the learned closure. Two correction strategies are investigated: direct learning, in which HF coefficients are predicted directly from LF inputs, and residual learning, which learns the discrepancy between LF and HF coefficients. The framework is demonstrated on a double shear layer and vortex merging problems governed by the two-dimensional Navier–Stokes equations. Results show that both correction strategies improve ROM accuracy over uncorrected ROM, with residual learning achieving consistently better performance than direct learning. Moreover, the two proposed deep generative model-based strategies provide uncertainty quantification for the corrected ROM coefficients, which is critical for assessing prediction confidence and supporting the reliable use of ROMs in practical applications.
\end{abstract}


\end{frontmatter}

\section{Introduction}\label{sec:intro}
High-fidelity (HF) simulations are necessary for accurately resolving the dynamics of complex multiscale systems, but their computational cost is often prohibitive, especially in many-query settings such as parametric studies, optimization, uncertainty propagation, and control. To reduce this burden, reduced-order models (ROMs) are commonly constructed by projecting the full-order dynamics onto a low-dimensional subspace defined by a small number of dominant basis functions \cite{liang2002proper, lucia2004reduced, padula2024brief}. Among the most widely used approaches, proper orthogonal decomposition (POD) provides an efficient reduced representation by approximating the full state field with a truncated expansion consisting of a mean field, a limited number of POD modes, and their associated time-dependent modal coefficients.

Despite their efficiency, POD-based ROMs may suffer from reduced predictive fidelity due to truncation of the POD expansion. In particular, the Galerkin POD (GPOD) framework evolves the retained modal coefficients by projecting the governing full-order operators onto the reduced POD basis \cite{ullmann2016pod}. However, this truncation neglects the influence of unresolved modes and discarded scales whose interactions with the resolved modes may remain dynamically important. As a result, the reduced system may fail to reproduce the true evolution of the retained coefficients, especially for systems exhibiting strongly-nonlinear dynamics.
This deficiency is commonly referred to as the closure problem in ROM \cite{ahmed2021closures}, and necessitates an additional correction, or closure term, to compensate for the missing interaction between resolved and unresolved scales. From this perspective, closure correction can be naturally interpreted as a multi-fidelity (MF) learning problem \cite{sanderse2024scientific, ahmed2021multifidelity}. Specifically, the output of the ROM is treated as a low-fidelity (LF) approximation, since it captures the dominant large-scale behavior of the system at a significantly reduced computational cost but does not fully account for the effects of truncated modes. In contrast, the corresponding coefficients obtained from the HF simulation are regarded as the HF target. The central objective is therefore to learn a correction operator that maps the LF information to HF-consistent coefficients, so that the corrected ROM can recover the accuracy of the full-order dynamics while retaining computational efficiency. 

Recent advances in operator learning, such as DeepONets \cite{lu2021learning, ahmed2023multifidelity} and Fourier neural operators (FNO) \cite{li2020fourier, lyu2023multi} have provided powerful tools for learning correction terms in reduced-order and MF modeling. Specifically, these methods are well suited for constructing mappings between LF and HF representations and have shown strong performance across a range of scientific computing problems \cite{lu2022comprehensive}. For example, \citet{ahmed2021multifidelity} formulated the closure modeling as an MF operator learning problem, where the GPOD model acts as the LF component and the learned operator estimates the HF correction needed to improve the evolution of the POD coefficients.

Despite their success, most existing correction frameworks based on DeepONet, FNO, or related neural architectures are fundamentally deterministic in the sense that they produce a single corrected output for a given LF input \cite{bulte2024probabilistic}. In many reduced-order settings, the mapping from LF coefficients to their HF counterparts is not strictly deterministic because the reduced representation does not fully preserve the information carried by unresolved modes. The numerical approximation errors and information loss induced by modal truncation can all introduce ambiguity into the correction process. As a result, the same LF state may correspond to multiple plausible HF-consistent corrections, particularly when the ROM is strongly under-resolved or when errors accumulate over long prediction horizons \cite{callaham2023multiscale, snyder2022reduced}.
Although deterministic correction models can provide a single best-fit prediction, they do not describe how uncertain the resulting prediction is. Without a probabilistic description of the LF-to-HF mapping, the corrected ROM may appear overly confident even when its prediction is inaccurate. Such overconfidence undermines the trustworthiness of the model and limits its usefulness in downstream tasks such as forecasting, control, and uncertainty propagation.

To address these challenges, we propose a deep generative MF closure framework with built-in uncertainty quantification based on conditional normalizing flows (CNFs). CNF learns flexible probability distributions by transforming a simple base distribution through a sequence of invertible transformations. In this framework, the normalizing flow models the conditional distribution of the HF coefficients, or alternatively the HF--LF discrepancy, given the LF ROM coefficients. As a result, the model can generate multiple plausible HF-consistent samples for the same LF input, from which predictive statistics such as the mean, variance, and credibility intervals can be estimated. In the current work, two correction strategies are investigated: direct learning, in which HF coefficients are predicted directly from LF inputs, and residual learning, in which the discrepancy between LF and HF coefficients is learned. We demonstrate this framework on two two-dimensional Navier-Stokes problems: a vortex merging problem double shear layer problem, and vortex merging problem. In particular, the vortex merging phenomenon has been key in studying a wide range of geophysical flows, including atmospheric storms and large oceanic eddies, and is increasingly used as a representative benchmark for ROM developments \cite{ahmed2023physics}. Our results demonstrate that both correction strategies (i.e., direct learning and residual learning) improve ROM accuracy compared to the uncorrected model, with residual learning consistently achieving better performance than direct learning. Furthermore, the probabilistic predictions provide uncertainty estimates that are crucial for assessing predictive confidence and enabling the reliable use of ROMs in practical applications (see also the recent work in \cite{bhola2025flow}).

The remainder of this paper is organized as follows. \Cref{sec:prob} introduces the problem formulation, including the governing partial differential equations for the full-order model (FOM) and its projection-based ROM formulation. \Cref{sec:method} presents the proposed MF framework, including the direct learning and residual learning strategies, together with the details of the CNF model. \Cref{sec:application} presents the application of the proposed framework to the aforementioned problems. Finally, conclusions and future research directions are discussed in \Cref{sec:conclusion}.

\section{Problem formulation}\label{sec:prob}

In this section we discuss the FOM initial and boundary value problem (IBVP), which represents the HF dynamical system of interest, and summarize the derivation of the GPOD ROM.

\subsection{Full-order model}
Consider a parametric nonlinear dynamical system defined on a spatial domain \(\Omega \subset \mathbb{R}^{d}\) over a time interval \(t \in [0,T]\). We consider the parametric IBVP of the form
\begin{align}
  \label{eq:fom_continuous}
  \frac{\partial u(\mathbm{x},t;\bm{\mu})}{\partial t} &= \mathcal{N}\!\left(u(\mathbm{x},t;\bm{\mu});\bm{\mu}\right), &&\quad \mathbm{x}\in\Omega,\; t\in(0,T],\\
  \label{eq:fom_bc}
  u(\mathbm{x},0;\bm{\mu}) &= u_{0}(\mathbm{x};\bm{\mu}), &&\quad \mathbm{x}\in\Omega,\\
  \label{eq:fom_ic}
  \mathcal{B}\!\left(u(\mathbm{x},t;\bm{\mu});\bm{\mu}\right) &= 0, &&\quad \mathbm{x}\in\partial\Omega,\; t\in(0,T],
\end{align}
where \(u(\mathbm{x},t;\bm{\mu})\) denotes the state variable, \(\bm{\mu}\in M\) is the vector of model parameters, \(M \subseteq \mathbb{R}^p \) is the space of model parameters, \(\mathcal{N}(\cdot)\) is a generally nonlinear differential operator representing the governing physics, and \(\mathcal{B}(\cdot)\) denotes the boundary conditions operator.

After a method-of-lines (MOL) discretization (via e.g. finite differences or finite elements), the IBVP of \Cref{eq:fom_continuous,eq:fom_bc,eq:fom_ic} is transformed into a system of first-order ordinary differential equations. Let \(N\) denote the number of spatial degrees of freedom and let \(\mathbm{u}(t;\bm{\mu}) \in \mathbb{R}^{N}\) denote the corresponding semi-discrete state vector. The MOL initial value problem (IVP) takes the form
\begin{equation}
  \dot{\mathbm{u}}(t;\bm{\mu})
  =\mathbm{f}\!\left(\mathbm{u}(t;\bm{\mu});\bm{\mu}\right),
  \quad
  \mathbm{u}(0;\bm{\mu})=\mathbm{u}_{0}(\bm{\mu}),
  \label{eq:fom_semidiscrete}
\end{equation}
where \(\mathbm{f}:\mathbb{R}^{N}\times M \rightarrow\mathbb{R}^{N}\)
collects the discrete linear and nonlinear operators arising from the spatial discretization. 

We refer to the IVP of \Cref{eq:fom_semidiscrete} as the FOM because it evolves in the original high-dimensional state space \(\mathbb{R}^{N}\). In many applications, \(N\) is very large and accurate time integration of the FOM requires substantial computational effort. This cost becomes particularly restrictive in multi-query settings where the governing model must be evaluated repeatedly for many instances of the parameters \(\bm{\mu}\).

\subsection{Galerkin POD reduced-order model} \label{sec:rom}

Let \(\mathbm{\bar{u}} \in \mathbb{R}^{N}\) denote a reference solution (e.g., the time-averaged velocity field). The full-order state \(\mathbm{u}(t;\bm{\mu})\) is then approximated by a truncated POD expansion of the form
\begin{equation}
  \label{eq:pod_expansion}
  \mathbm{u}(t;\bm{\mu})
  \approx
  \mathbm{\bar{u}} + \sum\nolimits_{i=1}^{R} a_i(t;\bm{\mu}) \mathbm{\phi}_i,
\end{equation}
where \(\mathbm{\phi}_i \in \mathbb{R}^{N}\), \(i=1,\dots,R\), are the retained POD basis vectors, \(a_i(t;\bm{\mu})\) are the corresponding time-dependent modal coefficients, and \(R \ll N\) is the reduced dimension. The quality of the approximation depends on the number of retained modes \(R\). In matrix form, the approximation can be written as
\begin{equation}
  \label{eq:pod_matrix_form}
  \mathbm{u}(t;\bm{\mu})
  \approx
  \mathbm{\bar{u}} + \mathbm{\Phi}\mathbm{a}(t;\bm{\mu}),
\end{equation}
where \(
\mathbm{\Phi} = [\mathbm{\phi}_1,\mathbm{\phi}_2,\dots,\mathbm{\phi}_R] \in \mathbb{R}^{N\times R}\) and \(\mathbm{a}(t;\bm{\mu}) = [a_1(t;\bm{\mu}),\dots,a_R(t;\bm{\mu})]^{\top} \in \mathbb{R}^{R}\).
The columns of \(\mathbm{\Phi}\) span the reduced subspace where the solution is approximated.

The POD basis can be constructed from snapshot data generated by the FOM. Suppose that \(N_s\) solution snapshots \(\mathbm{u}^{(1)},\mathbm{u}^{(2)},\dots,\mathbm{u}^{(N_s)} \in \mathbb{R}^{N}\) are collected at different times and/or parameter samples.
These snapshots are then assembled into the snapshot matrix \(\mathbm{X}=\begin{bmatrix} \mathbm{u}^{(1)} & \mathbm{u}^{(2)} & \cdots & \mathbm{u}^{(N_s)}
\end{bmatrix}\in \mathbb{R}^{N\times N_s}\).
The reference state is typically taken as the empirical mean of the snapshots, that is,
\begin{equation}
\mathbm{\bar{u}}
=\frac{1}{N_s}\sum\nolimits_{j=1}^{N_s}\mathbm{u}^{(j)}.
\label{eq:mean_state}
\end{equation}
Subtracting this mean from each snapshot yields the centered snapshot matrix
\begin{equation}
\mathbm{X}'
=\begin{bmatrix}
\mathbm{u}^{(1)}-\mathbm{\bar{u}} &
\mathbm{u}^{(2)}-\mathbm{\bar{u}} &
\cdots &
\mathbm{u}^{(N_s)}-\mathbm{\bar{u}}
\end{bmatrix}
\in \mathbb{R}^{N\times N_s}.
\label{eq:centered_snapshot_matrix}
\end{equation}

The POD modes are obtained by extracting the dominant orthogonal directions of
\(\mathbm{X}'\),
which can be readily achieved through the singular value decomposition
\begin{equation}
\mathbm{X}' = \mathbm{U}\mathbm{\Sigma}\mathbm{V}^{\top},
\label{eq:svd}
\end{equation}
where
\(\mathbm{U}\in\mathbb{R}^{N\times N}\)
contains the left singular vectors,
\(\mathbm{\Sigma}\in\mathbb{R}^{N\times N_s}\)
is the diagonal matrix of singular values, and
\(\mathbm{V}\in\mathbb{R}^{N_s\times N_s}\)
contains the right singular vectors. The POD basis vectors are chosen as the first \(R\)
dominant left singular vectors,
\begin{equation}
\mathbm{\Phi} = [\mathbm{\phi}_1,\mathbm{\phi}_2,\dots,\mathbm{\phi}_R]
=[\mathbm{U}_1,\mathbm{U}_2,\dots,\mathbm{U}_R],
\label{eq:pod_modes_svd}
\end{equation}
where \(\mathbm{U}_i\) denotes the \(i\)-th column of
\(\mathbm{U}\).
Equivalently, the POD modes may be obtained from the eigenvalue decomposition of the snapshot covariance matrix
\begin{equation}
\mathbm{X}'{\mathbm{X}'}^{\top}\mathbm{\phi}_i = \lambda_i \mathbm{\phi}_i,
\qquad i=1,\dots,R,
\label{eq:covariance_eig}
\end{equation}
with eigenvalues ordered such that \(\lambda_1 \ge \lambda_2 \ge \cdots \ge \lambda_R \ge \cdots \ge 0\).
The retained modes correspond to the largest eigenvalues and therefore capture the dominant energy content of the snapshot ensemble.
We quantify the retained energy content using the ``relative information criterion'' (RIC) metric, defined as
\begin{equation*}
  \text{RIC}(R) = 100 \times \frac{\sum^R_{i = 1} \sigma^2_i }{\sum^{\min \{ N, N_s \}}_{i = 1} \sigma^2_i },
\end{equation*}
where \(\sigma_i\) denotes the \(i\)th singular value of the snapshot matrix.

After constructing the POD basis functions, the burden is to compute the corresponding coefficients at any given time and/or parameter.
The POD basis vectors are orthonormal under the standard Euclidean inner product, that is,
\begin{equation}
\mathbm{\Phi}^{\top}\mathbm{\Phi} = \mathbm{I}_R,
\label{eq:orthonormality}
\end{equation}
where
\(\mathbm{I}_R\)
is the \(R\times R\) identity matrix. Consequently, for a given state \(\mathbm{u}(t;\bm{\mu})\),
the (true) reduced coefficients are obtained by orthogonal projection onto the POD subspace,
\begin{equation}
\mathbm{a}(t;\bm{\mu})
=\mathbm{\Phi}^{\top}\big(\mathbm{u}(t;\bm{\mu})-\mathbm{\bar{u}}\big).
\label{eq:projection_coeff}
\end{equation}
However, \Cref{eq:projection_coeff} relies on knowledge of the FOM solution $\mathbm{u}(t;\bm{\mu})$.
Instead, Galerkin projection can be applied to obtain reduced dynamics for which the full-order system is projected onto the POD subspace, leading to a GPOD ROM governing the time evolution of the reduced coefficients \(\mathbm{a}(t;\bm{\mu})\).
For the IVP of \Cref{eq:fom_semidiscrete}, the GPOD ROM takes the form
\begin{equation}
\dot{\mathbm{a}}(t;\bm{\mu})
=\mathbm{\Phi}^{\top}
\mathbm{f}\!\left(\mathbm{\bar{u}}+\mathbm{\Phi}\mathbm{a}(t;\bm{\mu});\bm{\mu}\right).
\label{eq:gpod_continuous}
\end{equation}
\Cref{eq:gpod_continuous} describes the reduced evolution entirely in the coefficient space
\(\mathbb{R}^{R}\),
and is referred to as the GPOD ROM. In other words, the original high-dimensional evolution of
\(\mathbm{u}(t;\bm{\mu}) \in \mathbb{R}^{N}\)
is replaced by the evolution of a much smaller coefficient vector
\(\mathbm{a}(t;\bm{\mu}) \in \mathbb{R}^{R}
\).

Let \(\mathbm{a}_{n}^{\mathrm{GPOD}}\) denote the reduced state at time \(t_n\). Then the one-step GPOD evolution can be written as
\begin{equation}
\mathbm{a}_{n+1}^{\mathrm{GPOD}}(\bm{\mu})
=\mathcal{G}\!\left(\mathbm{a}_{n}^{\mathrm{GPOD}}(\bm{\mu});\bm{\mu}\right),
\label{eq:gpod_discrete}
\end{equation}
where \(\mathcal{G}(\cdot;\bm{\mu})\)
denotes the one-step GPOD flow map that arises from applying an explicit time integration scheme to \Cref{eq:gpod_continuous}. The GPOD model evolves only the retained POD coefficients and is therefore much less expensive than the original FOM. 

\subsection{Closure modeling}

Although the GPOD system evolves the retained POD coefficients efficiently, it is not in general a closed representation of the full-order dynamics because the reduced approximation retains only the first \(R\) POD modes and omits the influence of the discarded modes. When \(R \ll N\), this truncation introduces a modeling error, and the resulting reduced system may deviate from the true evolution of the HF projected coefficients. This missing contribution is commonly referred to as the closure term, or, more broadly, the closure problem in ROM.

Let \(\mathbm{a}_{n}^{\mathrm{HF}}\) and 
\(\mathbm{a}_{n+1}^{\mathrm{HF}}\) denote the reduced coefficients obtained by projecting the HF solutions at time levels \(t_n\) and \(t_{n+1}\), respectively, onto the retained POD basis. Starting from the HF-projected coefficient \(\mathbm{a}_{n}^{\mathrm{HF}}\), the corresponding one-step GPOD prediction is given by \(\mathbm{a}_{n+1}^{\mathrm{GPOD}}(\bm{\mu}) = \mathcal{G}\!\left(\mathbm{a}_{n}^{\mathrm{HF}}(\bm{\mu});\bm{\mu}\right) \).
The discrepancy between the projected HF coefficients and the one-step GPOD prediction is then given by \(\mathbm{r}_{n+1} = \mathbm{a}_{n+1}^{\mathrm{HF}} - \mathbm{a}_{n+1}^{\mathrm{GPOD}} \). Equivalently, the closure relation can be written as
\begin{equation}
  \label{eq:closure_relation_discrete_map}
  \mathbm{a}_{n+1}^{\mathrm{HF}}(\bm{\mu})
  =\mathcal{G}\!\left(\mathbm{a}_{n}^{\mathrm{HF}}(\bm{\mu}) ;\bm{\mu}\right)
  +\mathbm{r}_{n+1}(\bm{\mu}),
\end{equation}
where the discrepancy \( \mathbm{r}_{n+1} \) serves the role of a closure correction term.
This form shows that the GPOD model provides only an incomplete LF propagation of the retained coefficients, and that an additional correction is required to recover the HF reduced dynamics.

The closure term may be interpreted in several equivalent ways. From a ROM perspective, it represents the influence of the discarded modes on the retained dynamics. From a model-correction perspective, it compensates for the error introduced by modal truncation. 
In the present work, this missing contribution is learned from data in a MF setting. Specifically, the GPOD evolution is treated as a LF prediction in the reduced space, while the projected HF coefficients provide the reference target. The objective is therefore not to replace the reduced model itself, but to learn a probabilistic closure correction that maps the GPOD prediction toward the corresponding HF reduced state while quantifying the uncertainty associated with this correction. 

\section{Methodology}\label{sec:method}

\begin{figure}[http]
  \centering
  \begin{subfigure}[b]{0.95\textwidth}
    \includegraphics[width=\textwidth]{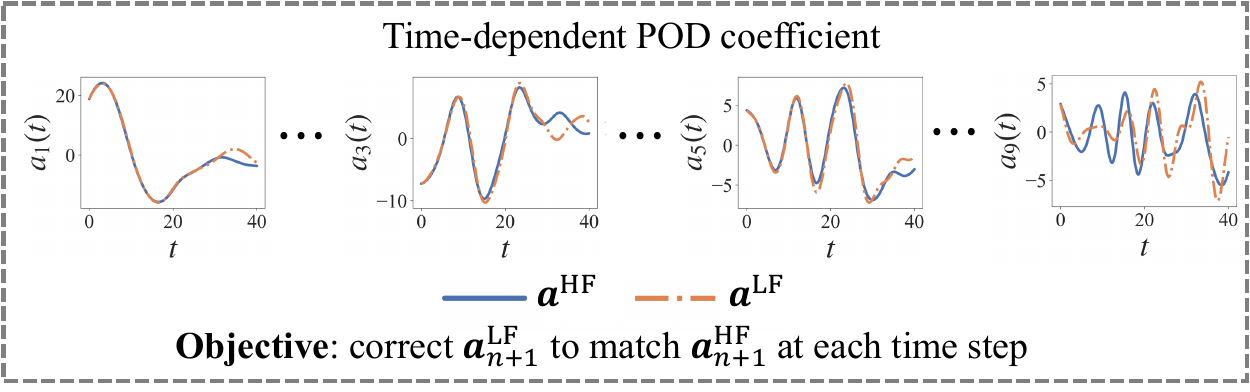}
    \caption{HF ROM state computed by projecting the FOM solution \(\mathbm{u}(t)\) into POD space, compared against the LF prediction computed by timestepping the ROM state.}
    \label{fig:HF_LF_errors}
  \end{subfigure}
  \vskip1ex
  \begin{subfigure}[b]{\textwidth}
    \includegraphics[width=\textwidth]{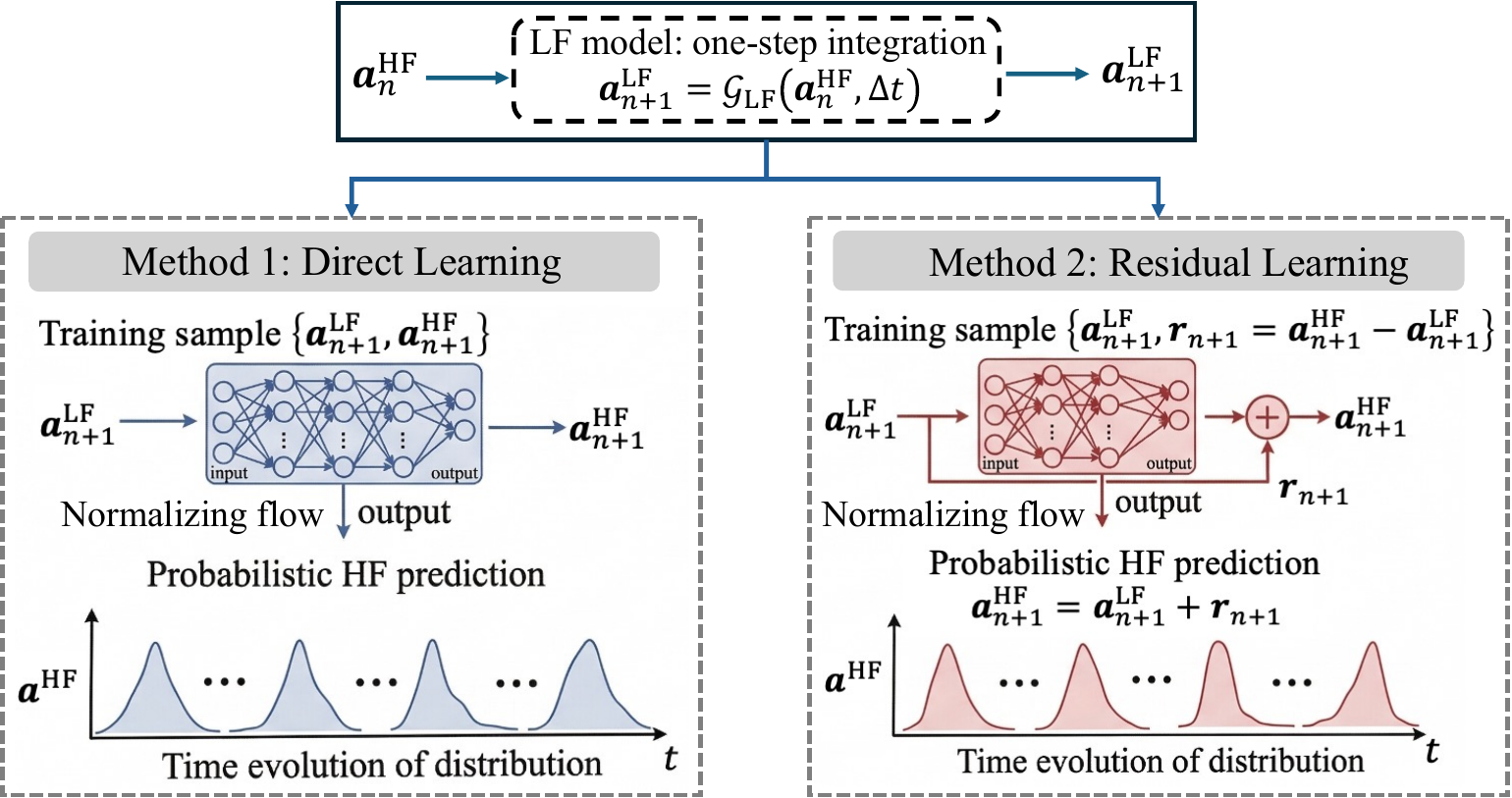}
    \caption{Probabilistic MF timestepping strategies.}
    \label{fig:pmf_timestepping}
  \end{subfigure}
  \caption{(a) Timestepping the ROM state using a LF model leads to prediction errors due to modal truncation and ROM approximations. (b) Proposed probabilistic multi-fidelity framework. HF predictions of the ROM state \(\mathbm{a}^{\mathrm{HF}}_{n+1}\) at time \(t_{n+1}\) are generated by (i) computing LF predictions \(\mathbm{a}^{\mathrm{LF}}_{n + 1}\) by timestepping the HF state \(\mathbm{a}^{\mathrm{HF}}_{n}\) from \(t_n\) to \(t_{n+1}\) using a LF model, and (ii) applying a probabilistic map \( \mathbm{a}^{\mathrm{LF}}_{n+1} \mapsto \mathbm{a}^{\mathrm{HF}}_{n+1} \). Two strategies for defining the probabilistic map are considered: Direct learning (DL) (\( \mathbm{a}^{\mathrm{LF}}_{n+1} \mapsto \mathbm{a}^{\mathrm{HF}}_{n+1} \)) and residual learning (RL) (\( \mathbm{a}^{\mathrm{LF}}_{n+1} \mapsto \mathbm{r}_{n+1} \), \(\mathbm{a}^{\mathrm{HF}}_{n+1} = \mathbm{a}^{\mathrm{LF}}_{n+1} + \mathbm{r}_{n+1}\)). Recursively applying this procedure leads to HF probabilistic timeseries predictions of the ROM state.} 
  \label{fig:workflow}
\end{figure}

In this section we present the proposed probabilistic MF closure-correction framework.
\Cref{fig:HF_LF_errors} shows the HF ROM state computed by projecting the FOM solution \(\mathbm{u}(t; \bm{\mu})\) of the vortex merger problem considered in \Cref{sec:vortex} for a certain value of \(\bm{\mu}\), and compares it against the LF ROM state obtained by timestepping the ROM state using a GPOD model.
It can be seen that there are noticeable discrepancies between the HF and LF POD coefficients, induced by modal truncation and reduced-model approximation, especially at high-index coefficients (e.g., $a_9(t)$).

Based on the closure relation \labelcref{eq:closure_relation_discrete_map}, the goal is to timestep the ROM state by combining LF timestepping and probabilistic MF corrections.
\Cref{fig:pmf_timestepping} illustrates the proposed timestepping probabilistic framework.
The HF ROM state at time \(t_{n+1}\) is simulated in two steps: First, we compute a LF prediction \(\mathbm{a}^{\mathrm{LF}}_{n + 1}(\bm{\mu})\) by timestepping the HF state \(\mathbm{a}^{\mathrm{HF}}_{n}(\bm{\mu})\) from \(t_n\) to \(t_{n+1}\) using the LF GPOD model.
Second, we apply a probabilistic correction to convert the LF prediction into the HF state \( \mathbm{a}^{\mathrm{HF}}_{n+1} \).
We explore two strategies for defining the probabilistic correction: In the direct learning (DL) approach, we model the probabilistic map \( \mathbm{a}^{\mathrm{LF}}_{n+1}(\bm{\mu}) \mapsto \mathbm{a}^{\mathrm{HF}}_{n+1}(\bm{\mu}) \) in terms of the conditional distribution \(p(\mathbm{a}^{\mathrm{LF}}_{n+1}(\bm{\mu}) \mid \mathbm{a}^{\mathrm{HF}}_{n+1}(\bm{\mu}))\).
In the residual learning (RL) approach, we instead model the probabilistic map \(\mathbm{a}^{\mathrm{LF}}_{n+1}(\bm{\mu}) \mapsto \mathbm{r}_{n+1}(\bm{\mu})\) via the conditional distribution \(p(\mathbm{r}_{n+1}(\bm{\mu}) \mid \mathbm{a}^{\mathrm{HF}}_{n+1}(\bm{\mu}))\) and compute the corrected HF as \(\mathbm{a}^{\mathrm{HF}}_{n+1}(\bm{\mu}) = \mathbm{a}^{\mathrm{LF}}_{n+1}(\bm{\mu}) + \mathbm{r}_{n+1}(\bm{\mu})\)).
Both strategies yield a time-evolving probabilistic prediction in ROM space rather than a single deterministic correction, thereby enabling explicit characterization of the uncertainty in the LF-to-HF mapping.

\subsection{Framework formulation and notation}\label{sec:notation}

We formulate the proposed methodology in the space of POD coefficients introduced in \Cref{sec:prob}.
For this purpose, we assume that the time domain is discretized into the sequence \( 0 < t_1 < \cdots t_{N_t - 1} < t_{N_t} \coloneqq T \), where \(N_t\) is the number of discrete time values, and that \(N_p\) values of \(\mathbm{\mu}\), \( \{ \bm{\mu}^{(k)} \in M \}^{N_p}_{k = 1}\), are available.
Furthermore, we denote by \(\mathcal{I}_t \coloneqq [1, N_t] \) and \( \mathcal{I}_p \coloneqq [1, N_p] \) the sets of time and parameter value indices.
Therefore, the cartesian product \( \mathcal{I}_s \coloneqq \mathcal{I}_t \times \mathcal{I}_p \) corresponds to the set of time snapshots available for training.
Finally, we introduce the multi-index notation \(k = (k_t, k_p)\), with \(k_t \in \mathcal{I}_t\) and \(k_p \in \mathcal{I}_p\), to denote the elements of \(\mathcal{I}_s\).

Additionally, we will employ the following input-target notation: Let \( \mathbm{x}, \mathbm{y} \in \mathbb{R}^R \) denote the input and target vectors, respectively.
We aim to approximate in a supervised manner the conditional distribution \(p(\mathbm{y} \mid \mathbm{x})\) by a parameterized distribution \( p_{\bm{\theta}}(\mathbm{y} \mid \mathbm{x}) \) with trainable parameters \(\mathbm{\theta}\) estimated using a dataset \(\mathcal{D} \coloneqq \{ (\mathbm{x}^{(k)}, \mathbm{y}^{(k)} ) \}_{k \in \mathcal{I}_s} \).
Furthermore, we assume that sampling from \(p_{\bm{\theta}}( \mathbm{y} \mid \mathbm{x})\) is performed as
\begin{equation}
  \label{eq:generic_transformation}
  \mathbm{y} \sim p_{\bm{\theta}}( \mathbm{y} \mid \mathbm{x}) \quad \Rightarrow \quad \mathbm{y} = \mathcal{M}_{\bm{\theta}}(\mathbm{x}; \mathbm{z}), \quad \mathbm{z} \sim \mathcal{N}(\mathbf{0}, \mathbf{I}_{N_z}),
\end{equation}
where \(\mathcal{M}_{\bm{\theta}} \colon \mathbb{R}^R \times \mathbb{R}^{N_z} \to \mathbb{R}^R\) is a parameterized transformation that transforms a latent code \(\mathbm{z} \in \mathbb{R}^{N_z}\) into the target \(\mathbm{y}\) using \(\mathbm{x}\) as a conditioning input.

\subsection{Direct learning formulation}\label{sec:DL}

We first consider the DL strategy, in which the probabilistic model is trained to learn the conditional distribution of the HF coefficients directly from their LF counterparts.
Specifically, for the \(k\)th data pair, the target \(\mathbm{y}^{(k)}\) corresponds to the HF coefficients \( \mathbm{a}^{\mathrm{HF}}_{k_t}(\bm{\mu}^{(k_p)}) \) obtained by projecting the FOM solution \(\mathbm{u}(t_{k_t}; \bm{\mu}^{(k_p)}) \) onto the truncated POD basis via \Cref{eq:projection_coeff}.
Similarly, the input \( \mathbm{x}^{(k)}\) corresponds to the LF coefficients \(\mathbm{a}_{k_t}^{\mathrm{LF}}(\bm{\mu}^{(k_p)})\) obtained by timestepping the GPOD equations \labelcref{eq:gpod_discrete} from \(t_{k_t - 1}\) to \(t_{k_t}\) for parameters \(\bm{\mu}^{(k_p)}\), that is, \(\mathbm{a}_{k_t}^{\mathrm{LF}}(\bm{\mu}^{(k_p)}) \coloneqq \mathcal{G}(\mathbf{a}^{\mathrm{HF}}_{k_t - 1}; \bm{\mu}^{(k_p)}) \).

Within this strategy, the learning task is to approximate the conditional distribution \(p(\mathbm{a}^{\mathrm{HF}}_{n + 1}(\bm{\mu}) \mid \mathbm{a}^{\mathrm{LF}}_{n + 1}(\bm{\mu}))\) via the parameterized distribution \(p_{\bm{\theta}}(\cdot \mid \cdot)\) with trainable parameters \(\bm{\theta}\) using the training dataset \(\mathcal{D}_{\mathrm{DL}} \coloneqq \{ (\mathbm{x}^{(k)}, \mathbm{y}^{(k)}) \}_{k \in \mathcal{I}_s} \), with \(\mathbm{x}^{(k)}, \mathbm{y}^{(k)} \) as defined above.
In contrast to a deterministic correction model, this approach assigns a conditional distribution to the HF coefficients as opposed to a single value.

We now describe how the DL strategy is embedded into the recursive MF prediction procedure. At inference time, timestepping from \(t_n\) to \(t_{n + 1}\) proceeds as follows:
Let \(\mathbm{a}_{n}^{\mathrm{DL}}(\bm{\mu})\) denote the corrected POD coefficient available at time \(t_n\).
We first compute the one-step LF prediction
\begin{equation*}
  \mathbm{a}_{n+1}^{\mathrm{LF}}(\bm{\mu}) = \mathcal{G}\!\left(\mathbm{a}_{n}^{\mathrm{DL}}(\bm{\mu}); \bm{\mu}\right).
\end{equation*}
The DL model then takes \(\mathbm{a}_{n+1}^{\mathrm{LF}}(\bm{\mu})\) as the conditioning input and returns a probabilistic prediction of the corresponding HF reduced coefficients, i.e., 
\begin{equation*}
  \mathbm{a}_{n+1}^{\mathrm{DL}}(\bm{\mu}) \sim p_{\bm{\theta}}\!\left( \mathbm{a}_{n+1}^{\mathrm{HF}}(\bm{\mu}) \mid \mathbm{a}_{n+1}^{\mathrm{LF}}(\bm{\mu}) \right).
\end{equation*}
That is, the MF prediction at each step is obtained in two stages: a LF propagation by the GPOD model, followed by a probabilistic HF correction by the learned DL.

This procedure can be applied recursively in time to produce an ensemble of predictions. Starting from an initial reduced state \(\mathbm{a}_{0}^{\mathrm{DL}}\), the \(m\)th sequence of MF predictions is computed by
\begin{equation}
  \label{eq:dl_recursive_sequence}
  \begin{aligned}
    \mathbm{a}_{n+1}^{\mathrm{LF},(m)}(\bm{\mu}) &= \mathcal{G}\!\left(\mathbm{a}_{n}^{\mathrm{DL},(m)} (\bm{\mu}) ; \bm{\mu} \right), \\
    \mathbm{a}_{n+1}^{\mathrm{DL},(m)}(\bm{\mu}) &= \mathcal{M}_{\bm{\theta}}\!\left( \mathbm{a}_{n+1}^{\mathrm{LF},(m)}(\bm{\mu}); \mathbm{z}^{(m)} \right),\\ \mathbm{z}^{(m)}_{n + 1} &\sim \mathcal{N}(\mathbf{0}, \mathbf{I}_{N_z}), \qquad n=0,1,\dots,T - 1
  \end{aligned}
\end{equation}
Applying this procedure \(N_e\) times leads to an ensemble \( \{\mathbm{a}_{1:N_t}^{\mathrm{DL},(m)} \}_{m=1}^{N_{\mathrm{e}}}\) of corrected MF trajectories.  forms an ensemble of corrected MF trajectories.  For example, with \(N_{\mathrm{e}}=200\), \(N_t=400\), and \(R=10\) retained POD modes, the resulting ensemble has size \(200 \times 400 \times 10\).
Each ensemble trajectory therefore represents one realization from the recursive predictive distribution induced by the composing the deterministic GPOD prediction model and the stochastic DL correction model.

Based on the ensemble of predicted trajectories, it is possible to estimate their ensemble mean
\begin{equation}
\bar{\mathbm{a}}_{n+1}^{\mathrm{DL}}
=
\frac{1}{N_{\mathrm{e}}}
\sum\nolimits_{m=1}^{N_{\mathrm{e}}}
\mathbm{a}_{n+1}^{\mathrm{DL},(m)}.
\label{eq:dl_mean}
\end{equation}
Predictive uncertainty is quantified in terms of the spread of the ensemble trajectories.
For each retained POD mode, the empirical standard deviation at time step \(n+1\) is
computed as
\begin{equation}
\hat{\sigma}_{n+1,j}^{\mathrm{DL}}
=
\left[
\frac{1}{N_{\mathrm{e}}-1}
\sum\nolimits_{m=1}^{N_{\mathrm{e}}}
\left(
a_{n+1,j}^{\mathrm{DL},(m)}
-
\bar{a}_{n+1,j}^{\mathrm{DL}}
\right)^2
\right]^{1/2},
\quad
j=1,\dots,R ,
\label{eq:dl_std_component}
\end{equation}
where \(a_{n+1,j}^{\mathrm{DL},(m)}\) and
\(\bar{a}_{n+1,j}^{\mathrm{DL}}\) denote the \(j\)-th components of
\(\mathbm{a}_{n+1}^{\mathrm{DL},(m)}\) and
\(\bar{\mathbm{a}}_{n+1}^{\mathrm{DL}}\), respectively.

\subsection{Residual learning formulation}\label{sec:RL}

We next consider the RL strategy, in which the probabilistic model is trained to learn the distribution of the residual \( \mathbm{r}_{n+1} \) conditional on the LF coefficients.
Similarly to the DL strategy, for the \(k\)th data pair we denote by \(\mathbm{a}^{\mathrm{HF}}_{k_t}(\bm{\mu}^{(k_p)})\) the HF coefficients obtained by projecting the FOM solution \(\mathbm{u}(t_{k_t}; \bm{\mu}^{(k_p)}) \) onto the truncated POD basis via \Cref{eq:projection_coeff}, and by \(\mathbm{a}_{k_t}^{\mathrm{LF}}(\bm{\mu}^{(k_p)}) \coloneqq \mathcal{G}(\mathbf{a}^{\mathrm{HF}}_{k_t - 1}; \bm{\mu}^{(k_p)}) \) the LF coefficients obtained by timestepping the GPOD equations \labelcref{eq:gpod_discrete} from \(t_{k_t - 1}\) to \(t_{k_t}\) for parameters \(\bm{\mu}^{(k_p)}\).
The target \(\mathbm{y}^{(k)}\) is taken to be the vector of residuals \( \mathbm{r}_{k_t}(\bm{\mu}^{(k_p)}) \coloneqq \mathbm{a}^{\mathrm{HF}}_{k_t}(\bm{\mu}^{(k_p)}) - \mathbm{a}^{\mathrm{LF}}_{k_t}(\bm{\mu}^{(k_p)}) \), and the input \( \mathbm{x}^{(k)}\) is taken to be \(\mathbm{a}_{k_t}^{\mathrm{LF}}(\bm{\mu}^{(k_p)})\).
The learning task is then to approximate the conditional distribution \(p(\mathbm{r}_{n + 1}(\bm{\mu}) \mid \mathbm{a}^{\mathrm{LF}}_{n + 1}(\bm{\mu}) )\) via the parameterized distribution \(p_{\bm{\theta}}(\cdot \mid \cdot)\) with trainable parameters \(\bm{\theta}\) using the training dataset \(\mathcal{D}_{\mathrm{DL}} \coloneqq \{ (\mathbm{x}^{(k)}, \mathbm{y}^{(k)}) \}_{k \in \mathcal{I}_s} \), with inputs and outputs \(\mathbm{x}^{(k)}, \mathbm{y}^{(k)} \) as defined above.

The RL strategy is embedded into the recursive MF prediction procedure as follows. Starting from an initial reduced state \(\mathbm{a}_{0}^{\mathrm{RL}}\), the \(m\)th sequence of MF predictions is computed by
\begin{equation}
  \label{eq:rl_recursive_sequence}
  \begin{aligned}
    \mathbm{a}_{n+1}^{\mathrm{LF},(m)}(\bm{\mu}) &= \mathcal{G}\!\left(\mathbm{a}_{n}^{\mathrm{RL},(m)} (\bm{\mu}) ; \bm{\mu} \right), \\
    \mathbm{r}_{n+1}^{(m)}(\bm{\mu}) &= \mathcal{M}_{\bm{\theta}}\!\left( \mathbm{a}_{n+1}^{\mathrm{LF},(m)}(\bm{\mu}); \mathbm{z}^{(m)} \right),\\
    \mathbm{z}^{(m)}_{n + 1} &\sim \mathcal{N}(\mathbf{0}, \mathbf{I}_{N_z}),\\
    \mathbm{a}_{n+1}^{\mathrm{RL}, (m)}(\bm{\mu}) &= \mathbm{a}_{n+1}^{\mathrm{LF}, (m)}(\bm{\mu}) + \mathbm{r}_{n+1}^{(m)}(\bm{\mu}), \quad n=0,1,\dots,T - 1. \\
  \end{aligned}
\end{equation}
Based on the ensemble of RL samples, the corrected mean prediction and the predictive uncertainty are evaluated in the same manner as in the DL strategy using expressions analogous to \Cref{eq:dl_mean,eq:dl_std_component}.

\subsection{Conditional normalizing flow}\label{sec:CNF}

\begin{figure}[http]
  \centering
  \includegraphics[width=\textwidth]{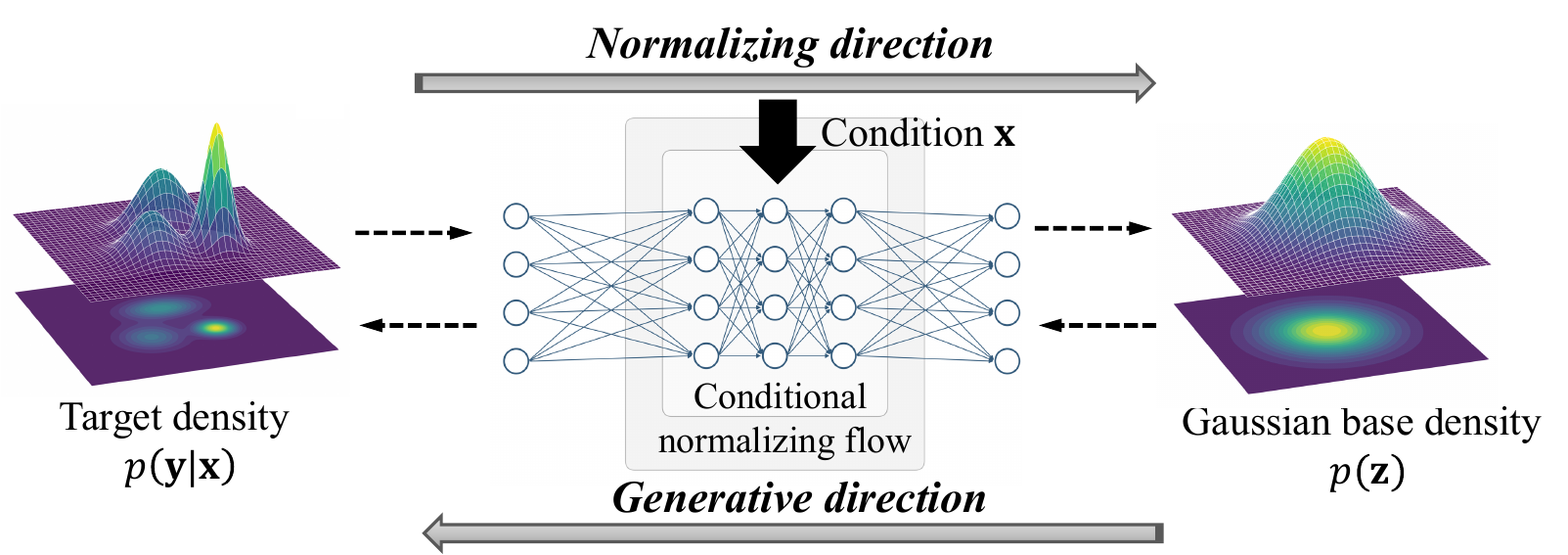}
  \caption{Density modeling via conditional normalizing flows (CNFs).
    In the generative direction, the complex conditional target density of outputs \(\mathbm{y} \mid \mathbm{x}\) is represented by transforming a Gaussian base density in the space of latent codes \(\mathbm{z}\) using a conditional invertible transformation.
  In the normalizing direction, the distribution of outputs is transformed into the distribution of latent codes.}
  \label{fig:cnf_schematic}
\end{figure}

In the present work we employ CNFs \cite{papamakarios2021normalizing,zeng2025solving,zeng2023recursive,zeng2025model} to model the conditional probability distributions of the DL and RL strategies.
Using the unified notation introduced in \Cref{sec:notation}, CNFs provide the conditional distribution model \(p_{\bm{\theta}}\!\left( \mathbm{y} \, | \, \mathbm{x} \right)\) with trainable parameters \(\bm{\theta}\).
\Cref{fig:cnf_schematic} illustrates the process of density modeling with CNFs: The conditioning input \(\mathbm{x}\) is supplied as auxiliary information, and the output \(\mathbm{y}\) is estimated by passing a vector of latent codes \(\mathbm{z} \in \mathbb{R}^R\) with a simple distribution through a bijective transformation.
The normalizing direction maps the targets to latent codes, whereas the generative direction maps latent codes back to targets for prediction and uncertainty quantification.
As specified in \Cref{sec:notation}, we require the latent codes to follow a standard multivariate Gaussian distribution, that is \(\mathbm{z} \sim \mathcal{N}(\mathbm{0},\mathbf{I}_R)\).

More specifically, CNFs are defined by the invertible transformation conditioned on \(\mathbm{x}\)
\begin{equation}
\mathbm{z}
=\mathcal{T}_{\bm{\theta}}(\mathbm{y};\mathbm{x}),
\quad
\mathbm{y}
=\mathcal{T}_{\bm{\theta}}^{-1}(\mathbm{z};\mathbm{x}).
\label{eq:cnf_bijection}
\end{equation}
In practice, this transformation is typically constructed as a composition of multiple invertible blocks. Let
\(\mathbm{h}_{0} \coloneqq \mathbm{y}\) and \(\mathbm{h}_{K} \coloneqq \mathbm{z}\).
We then introduce the sequence of intermediate transformations
\begin{equation}
  \mathbm{h}_{k}
  \coloneqq \mathcal{T}_{k,\bm{\theta}}(\mathbm{h}_{k-1};\mathbm{x}), \quad k=1,\dots,K,
\label{eq:cnf_composition_forward}
\end{equation}
such that the (forward) normalizing transformation is given by
\begin{equation}
  \mathbm{z}
  =\mathcal{T}_{K,\bm{\theta}} \circ \mathcal{T}_{K-1,\bm{\theta}} \circ \cdots \circ \mathcal{T}_{1,\bm{\theta}}(\mathbm{y};\mathbm{x}).
\label{eq:cnf_composition_compact}
\end{equation}
The (inverse) generative mapping is then obtained by reversing these transformations.
By the change-of-variables formula, the conditional density of \(\mathbm{y}\) given \(\mathbm{x}\) is \cite{radev2020bayesflow}
\begin{equation}
p_{\bm{\theta}}\!\left(
\mathbm{y} \, | \, \mathbm{x}
\right)
=p_{\mathbm{z}}\!\left(
\mathcal{T}_{\bm{\theta}}(\mathbm{y};\mathbm{x})
\right)
\left|
\det
\left(
\frac{\partial \mathcal{T}_{\bm{\theta}}(\mathbm{y};\mathbm{x})}{\partial \mathbm{y}}
\right)
\right|,
\label{eq:cnf_change_of_variables}
\end{equation}
where $p_{\mathbm{z}}$ denotes the density of the standard Gaussian base distribution, and \( \partial \mathcal{T}_{\bm{\theta}}(\mathbm{y};\mathbm{x}) / \partial \mathbm{y} \) denotes the Jacobian matrix of the transformation $\mathcal{T}_{\bm{\theta}}(\mathbm{y};\mathbm{x})$ with respect to $\mathbm{y}$.
Equivalently, in logarithmic form,
\begin{equation}
\log p_{\bm{\theta}}\!\left(
\mathbm{y} \, | \, \mathbm{x}
\right)
=\log p_{\mathbm{z}}(\mathbm{z})+
\sum\nolimits_{k=1}^{K}\log
\left|
\det
\left(
\frac{\partial \mathbm{h}_{k}}{\partial \mathbm{h}_{k-1}}
\right)
\right|.
\label{eq:cnf_log_density}
\end{equation}
In practice, the invertible blocks are carefully constructed so that the Jacobian determinants can be evaluated efficiently, which is essential for scalable training.

Accordingly, the CNF model parameters \(\bm{\theta}\) are estimated via conditional maximum likelihood estimation. Given a training dataset
\(\mathcal{D}
=\left\{
(\mathbm{x}^{(i)}, \mathbm{y}^{(i)})
\right\}_{i=1}^{N_s},\)
the corresponding empirical maximum-likelihood estimator is defined as
\begin{equation}
\bm{\theta}^{*}
\coloneqq \arg\min_{\bm{\theta}}
\mathcal{L}_{\mathrm{CNF}}(\bm{\theta}),
\label{eq:cnf_optimal_theta}
\end{equation}
with ensemble negative log-likelihood
\begin{equation}
\mathcal{L}_{\mathrm{CNF}}(\bm{\theta})
=-\frac{1}{N_s}
\sum_{i=1}\nolimits^{N_s}
\log p_{\bm{\theta}}\!\left(
\mathbm{y}^{(i)} \, | \, \mathbm{x}^{(i)}
\right).
\label{eq:cnf_training_objective}
\end{equation}
Using the change-of-variables relation in \Cref{eq:cnf_change_of_variables}, this loss function can be written explicitly as
\begin{equation}
\mathcal{L}_{\mathrm{CNF}}(\bm{\theta})
=-\frac{1}{N_s}
\sum\nolimits_{i=1}^{N_s}
\left[
\log p_{\mathbm{z}}(\mathbm{z}^{(i)})
+
\log
\left|
\det
\left(
\frac{\partial \mathcal{T}_{\bm{\theta}}(\mathbm{y}^{(i)};\mathbm{x}^{(i)})}{\partial \mathbm{y}^{(i)}}
\right)
\right|
\right],
\label{eq:cnf_explicit_loss}
\end{equation}
with \(\mathbm{z}^{(i)} = \mathcal{T}_{\bm{\theta}}(\mathbm{y}^{(i)};\mathbm{x}^{(i)})\).
Due to the fact that the latent codes
\(\mathbm{z}\)
follows the standard Gaussian distribution, their log-density can be written as
\begin{equation}
\log p_{\mathbm{Z}}(\mathbm{z})
=-\frac{R}{2}\log(2\pi)
-\frac{1}{2}\|\mathbm{z}\|_{2}^{2}.
\label{eq:cnf_gaussian_logdensity}
\end{equation}
Since the term \(R \log(2\pi) / 2\)
is constant with respect to \(\bm{\theta}\),
it does not affect the optimization and may be omitted. Substituting \Cref{eq:cnf_gaussian_logdensity} into \Cref{eq:cnf_explicit_loss} yields the loss formula
\begin{equation}
\mathcal{L}_{\mathrm{CNF}}(\bm{\theta})
=\frac{1}{N_s}
\sum\nolimits_{i=1}^{N_s}
\left[
\frac{1}{2}\|\mathbm{z}^{(i)}\|_{2}^{2}
-\log
\left|
\det
\left(
\frac{\partial \mathcal{T}_{\bm{\theta}}(\mathbm{y}^{(i)};\mathbm{x}^{(i)})}{\partial \mathbm{y}^{(i)}}
\right)
\right|
\right].
\label{eq:cnf_training_objective_simplified}
\end{equation}
The minimization can be carried out using any stochastic gradient-based optimization algorithm.

In the generative direction, the CNF maps a latent sample \(\mathbm{z}^{(m)} \sim \mathcal{N}(\mathbm{0},\mathbf{I}_R)\) to the target space for a given conditioning input \(\mathbm{x}\).
The \(m\)th sample of the target distribution is then computed as
\begin{equation}
\mathbm{y}^{(m)}
=\mathcal{T}_{\bm{\theta}}^{-1}(\mathbm{z}^{(m)};\mathbm{x}),
\quad
\mathbm{z}^{(m)} \sim \mathcal{N}(\mathbm{0},\mathbf{I}),
\quad
m=1,\dots,M.
\label{eq:cnf_sampling_general}
\end{equation}
Repeated sampling from the latent Gaussian distribution therefore yields an ensemble of conditional predictions in target space. This ensemble provides both a mean prediction and an uncertainty estimate through the sample spread.

\subsection{Masked autoregressive flow}\label{sec:MAF}

We employ masked autoregressive flows (MAFs) \cite{papamakarios2017masked} to construct the normalizing invertible transformation.
MAFs construct this transformation as a composition of \(K\) autoregressive invertible blocks.
The \(k\)th MAF block is the autoregressive transformation defined componentwise by
\begin{equation}
h_{k,j}=\left(h_{k-1,j}-\mu_{k,j}(\mathbm{h}_{k-1,1:j-1};\mathbm{x})
\right)
\exp\!\left(
-\alpha_{k,j}(\mathbm{h}_{k-1,1:j-1};\mathbm{x})
\right),
\quad
j=1,\dots,R,
\label{eq:maf_forward_component}
\end{equation}
where \(\mu_{k,j}(\cdot;\mathbm{x})\) and \(\alpha_{k,j}(\cdot;\mathbm{x})\) denote shift and log-scale functions, respectively, and
\(\mathbm{h}_{k-1,1:j-1}\) collects the first \(j-1\) components of \(\mathbm{h}_{k-1}\).
Since the transformation of the \(j\)th component depends only on the preceding components and the condition \(\mathbm{x}\),
\Cref{eq:maf_forward_component} defines an autoregressive map.
The corresponding inverse transformation, used in the generative direction, is given by
\begin{equation}
h_{k-1,j}=\mu_{k,j}(\mathbm{h}_{k-1,1:j-1};\mathbm{x})+
\exp\!\left(
\alpha_{k,j}(\mathbm{h}_{k-1,1:j-1};\mathbm{x})
\right)
h_{k,j},
\quad
j=1,\dots,R.
\label{eq:maf_inverse_component}
\end{equation}
Because \(\mu_{k,j}\) and \(\alpha_{k,j}\) depend on previously generated components, the inverse map is evaluated sequentially from \(j=1\) to \(R\).

A key advantage of MAFs is that the Jacobian matrix of the forward transformation in \Cref{eq:maf_forward_component} is triangular. Specifically, the Jacobian of the \(k\)th block in \Cref{eq:cnf_log_density} is
\begin{equation}
\mathbm{J}_{k}=\frac{\partial \mathbm{h}_{k}}{\partial \mathbm{h}_{k-1}}=
\begin{bmatrix}
\exp(-\alpha_{k,1}) & 0 & 0 & \cdots & 0 \\
\ast & \exp(-\alpha_{k,2}) & 0 & \cdots & 0 \\
\ast & \ast & \exp(-\alpha_{k,3}) & \cdots & 0 \\
\vdots & \vdots & \vdots & \ddots & \vdots \\
\ast & \ast & \ast & \cdots & \exp(-\alpha_{k,R})
\end{bmatrix},
\label{eq:maf_jacobian}
\end{equation}
where the symbol \(\ast\) denotes nonzero entries arising from the dependence of \(\mu_{k,j}\)
and \(\alpha_{k,j}\) on the preceding components. Since
\(\mathbm{J}_{k}\) is lower triangular, its determinant is simply the product of its diagonal entries, that is,
\begin{equation}
\det(\mathbm{J}_{k})
=\prod_{j=1}^{R}
\exp(-\alpha_{k,j}),
\quad
\log
\left|
\det(\mathbm{J}_{k})
\right|
=-\sum_{j=1}\nolimits^{R}\alpha_{k,j}.
\label{eq:maf_logdet}
\end{equation}
This closed-form expression makes exact likelihood evaluation computationally efficient.

In this work, each autoregressive block in the MAF is implemented using a masked autoencoder for distribution estimation (MADE) \cite{germain2015made}. The role of the MADE network is to output the shift and log-scale parameters (\(\mu_{k,j}\) and \(\alpha_{k,j}\)) for all components \(j=1,\dots,R\) in a single forward pass while preserving the autoregressive dependency required by \Cref{eq:maf_forward_component}.
By stacking multiple MADE-parameterized autoregressive blocks, the resulting MAF model provides a flexible conditional density estimator with exact likelihood evaluation and efficient optimization.

\section{Applications to vortical flow problems}\label{sec:application}

In this section we apply the proposed MF framework to two canonical vortical flow problems: a double shear layer problem and a vortex merger problem. These examples are selected because they exhibit nonlinear flow evolution, strong modal interactions, and pronounced discrepancies between LF and HF reduced-order representations, providing representative test cases for assessing the capability of the proposed probabilistic MF framework to correct LF dynamics, recover HF modal evolution, and quantify predictive uncertainty.

\subsection{A double shear layer problem}\label{sec:double_shear}

A two-dimensional double shear layer problem is first considered to validate the proposed MF framework. This problem describes the nonlinear evolution of two oppositely oriented shear layers subject to a small perturbation, which leads to vortex roll-up, vortex interaction, and complex coherent flow structures \cite{bell1989second}. It provides a challenging benchmark for reduced-order modeling because the low-dimensional dynamics can be strongly nonlinear and sensitive to the Reynolds number \cite{ahmed2019memory}.

The flow is governed by the dimensionless 2D Navier--Stokes equation in the vorticity--streamfunction formulation,
\begin{equation}
    \dfrac{\partial \omega}{\partial t}  + J(\omega, \psi) 
    = \dfrac{1}{\mathrm{Re}} \nabla^2 \omega,
    \label{eq:dsl_ns}
\end{equation}
where $\omega$ and $\psi$ denote the vorticity and streamfunction, respectively. Here, $\text{Re}$ denotes the Reynolds number, which represents the ratio of inertial forces to viscous forces. The Jacobian operator, $J(\cdot,\cdot)$, and Laplacian $\nabla^2(\cdot)$ are defined
\begin{align}
    J(\omega,\psi)  &= \dfrac{\partial \omega}{\partial x} \dfrac{\partial \psi}{\partial y} - \dfrac{\partial \omega}{\partial y} \dfrac{\partial \psi}{\partial x}, \\
    \nabla^2 \omega &= \dfrac{\partial^2 \omega}{\partial x^2} + \dfrac{\partial^2 \omega}{\partial y^2}.
\end{align}
Finally, both the vorticity and streamfunction are linked through Poisson equation
\begin{equation}
    \nabla^2 \psi = - \omega. \label{eq:poisson}
\end{equation}

For the double shear layer setup, we consider a periodic square domain $(x,y)\in[0,2\pi]\times[0,2\pi]$. The initial velocity field is prescribed as
\begin{equation}
u(x,y,0)=
\begin{cases}
\tanh\big(\rho(y-\pi/2)\big), & y \leq \pi,\\
\tanh\big(\rho(3\pi/2-y)\big), & y > \pi,
\end{cases}
\qquad
v(x,y,0)=\epsilon \sin(x),
\label{eq:dsl_initial_velocity}
\end{equation}
where $\rho$ controls the thickness of the shear layers and $\epsilon$ denotes the amplitude of the initial perturbation. The corresponding initial vorticity field is obtained from
$\omega=\partial v/\partial x-\partial u/\partial y$ as follows:
\begin{equation}
    \omega(x,y,0) = 
    \begin{cases}
    \epsilon \cos(x) - \rho \cosh^{-2}\big(\rho(y-\pi/2)\big), & y \leq \pi,\\
    \epsilon \cos(x) + \rho \cosh^{-2}\big(\rho(3\pi/2-y)\big), & y > \pi.
    \end{cases}
\end{equation}

\subsubsection{High-fidelity coefficients: snapshot projection}
\begin{figure}[H]
  \centering
  \begin{subfigure}[b]{0.48\textwidth}
    \centering
    \includegraphics[width=\textwidth]{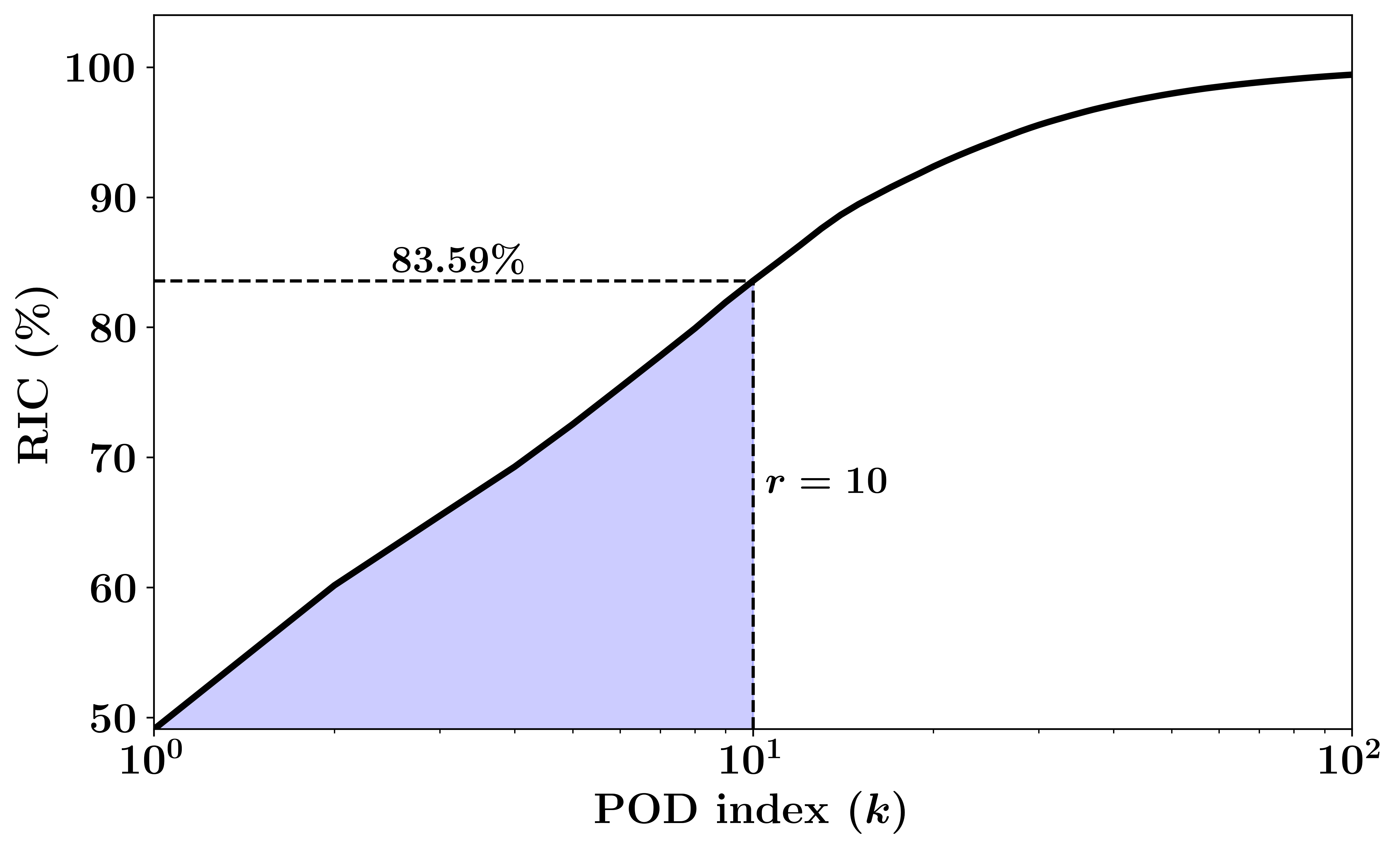}
    \caption{Double shear layer problem.}
    \label{fig:RIC_double_shear}
  \end{subfigure}
  \hfill
  \begin{subfigure}[b]{0.48\textwidth}
    \centering
    \includegraphics[width=\textwidth]{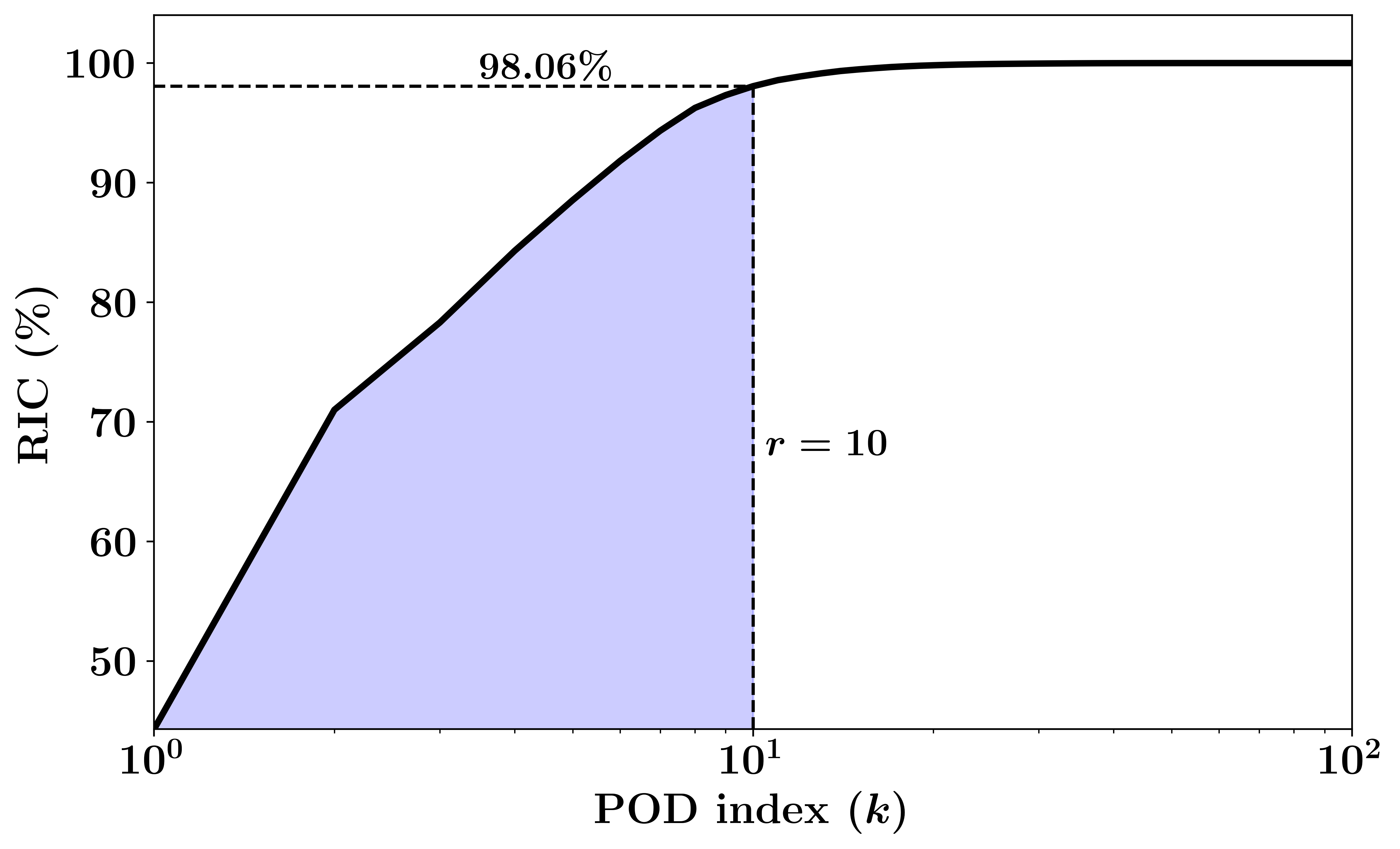}
    \caption{Vortex merger problem.}
    \label{fig:RIC_vortex_merger}
  \end{subfigure}
  \caption{Relative information content as a function of the number of retained POD modesl for (a) the double shear layer problem, and (b) the vortex merger problem.}
  \label{fig:RIC}
\end{figure}

To generate the HF solution, the FOM is solved using a third-order Arakawa scheme \cite{arakawa1966computational} for spatial discretization and a third-order total variation diminishing Runge--Kutta scheme (TVD-RK3) \cite{gottlieb1998total} for temporal integration  on a Cartesian grid of $256 \times 256$ points in the $x$- and $y$-directions. The initial condition parameters are set as $\rho=15/\pi$ and $\epsilon=0.05$. $401$ vorticity snapshots are collected over the time interval $t \in [0,40]$ for different Reynolds numbers. Global POD basis functions of the vorticity field, denoted by $\bm{\Phi}^{\omega}$, are constructed from the collected vorticity snapshots, as described in \Cref{sec:rom}. The corresponding streamfunction basis functions $\bm{\Phi}^{\psi}$ are then computed by solving the associated Poisson equation,
\begin{equation}
    \nabla^2 \bm{\Phi}^{\psi} = - \bm{\Phi}^{\omega}.
    \label{eq:dsl_poisson_pod}
\end{equation}
Although this treatment does not guarantee that $\bm{\Phi}^{\psi}$ is orthogonal, it preserves the kinematic relationship between vorticity and streamfunction and allows the same temporal coefficients to be used for both variables:
\begin{align}
    \omega(x,y,t;\mathrm{Re}) 
    &\approx \bar{\omega}(x,y;\mathrm{Re}) 
    + \sum_{i=1}^{R} a_i(t;\mathrm{Re}) \phi_i^{\omega}(x,y),
    \label{eq:dsl_wpod} \\
    \psi(x,y,t;\mathrm{Re}) 
    &\approx \bar{\psi}(x,y;\mathrm{Re}) 
    + \sum_{i=1}^{R} a_i(t;\mathrm{Re}) \phi_i^{\psi}(x,y).
    \label{eq:dsl_spod}
\end{align}
The HF POD coefficients are then obtained by projecting the full-order vorticity field onto the retained vorticity basis.
For this problem we retain 10 POD modes, corresponding to an RIC of \(\sim 84 \%\).
RIC as a function of the number of retained modes is shown in \Cref{fig:RIC_double_shear}.

\subsubsection{Low-fidelity model: Galerkin POD}
The LF model is constructed by substituting the POD approximations of \Cref{eq:dsl_wpod,eq:dsl_spod} into the vorticity--streamfunction equation and performing Galerkin projection onto the vorticity basis functions. This leads to a reduced-order dynamical system for the POD coefficients,
\begin{equation}
\dfrac{\mathrm{d} a_k}{\mathrm{d}t}
= \mathcal{C}_k
+ \sum_{i=1}^{R}\mathcal{L}_{ki} a_i
+ \sum_{i=1}^{R}\sum_{j=1}^{R} \mathcal{N}_{kij} a_i a_j,
\label{eq:dsl_gpod}
\end{equation}
where $\mathcal{C}_k$, $\mathcal{L}_{ki}$, and $\mathcal{N}_{kij}$ denote the constant, linear, and nonlinear Galerkin projection terms, respectively, given by
\begin{equation}
    \begin{aligned}
        \mathcal{C}_{k} &= \bigg( \phi^{\omega}_k , -J(\bar{\omega},\bar{\psi}) + \dfrac{1}{\text{Re}} \nabla^2 \bar{\omega} \bigg), \\
        \mathcal{L}_{ik} &= \bigg( \phi^{\omega}_k , -J(\bar{\omega},\phi^{\psi}_j)  -J(\phi^{\omega}_j,\bar{\psi}) + \dfrac{1}{\text{Re}} \nabla^2 \phi^{\omega}_j  \bigg), \\
        \mathcal{N}_{ijk} &= \bigg( \phi^{\omega}_k , -J(\phi^{\omega}_j,\phi^{\psi}_k) \bigg),
    \end{aligned} \label{eq:gpod_terms}
\end{equation}
with $\big(\cdot,\cdot \big)$ denoting the inner product. The form of \Cref{eq:dsl_gpod,eq:gpod_terms} takes advantage of the polynomial nonlinearity in governing equations to precompute the GPOD model terms during the offline stage. Applying explicit time integration schemes (e.g., Runge-Kutta family) to \cref{eq:dsl_gpod} defines the GPOD one-timestep mapping $\mathcal{G}$, and the LF datasets can be generated accordingly.

\subsubsection{CNF model training}
\label{sec:double_CNF_train}
For the double shear layer problem, the flow dynamics are represented in the reduced POD space using the first 10 time-dependent modal coefficients. The complete dataset contains 10 Reynolds-number cases, with each case consisting of 401 temporal snapshots over \(t \in [0,40]\). Thus, the complete LF and HF trajectories can be organized as tensors of size \(10 \times 401 \times 10\), corresponding to the Reynolds-number index, time index, and POD-mode index, respectively. Only the first 381 snapshots for each Reynolds-number case, corresponding to the interval \(t \in [0, 38]\), are used for training.
The trained closure model is subsequently evaluated through probabilistic recursive sequential prediction over the complete interval \(t \in [0,40]\). Therefore, the final 20 snapshots, corresponding to the terminal interval  \(38 < t \leq 40\), are employed to evaluate temporal extrapolation.

The probabilistic closure correction is learned using a CNF model implemented with the MAF architecture described in \Cref{sec:MAF}. The MAF consists of five MADE-based autoregressive transformations, with each transformation defined in the 10-dimensional POD coefficient space. Two hidden layers with 64 neurons are used in each autoregressive block. The model parameters are optimized using Adam with a learning rate of \(10^{-4}\), a batch size of 128, and 50,000 training iterations.

\subsubsection{Closure correction results}\label{sec:double_results}

After the CNF model is trained, we conduct probabilistic sequential prediction over the complete time horizon \(t\in[0,40])\).
Rollout is initialized using the corresponding HF POD coefficients at the first temporal snapshot, after which the POD states are generated recursively in time.
Because no snapshots over the interval \(38 < t \le 40)\) are used during model training, the final two time units constitute a temporal extrapolation interval for the learned closure correction.
To characterize the predictive uncertainty, the sequential rollout procedure is repeated 200 times using different random seeds, producing an ensemble of 200 probabilistic trajectories. It is important to note that, within each individual trajectory, only one CNF sample is drawn at each time step, providing an ensemble-based approximation of the stochastic temporal evolution of the POD coefficients.

\begin{figure}[h]
  \centering
  \includegraphics[width=\textwidth]{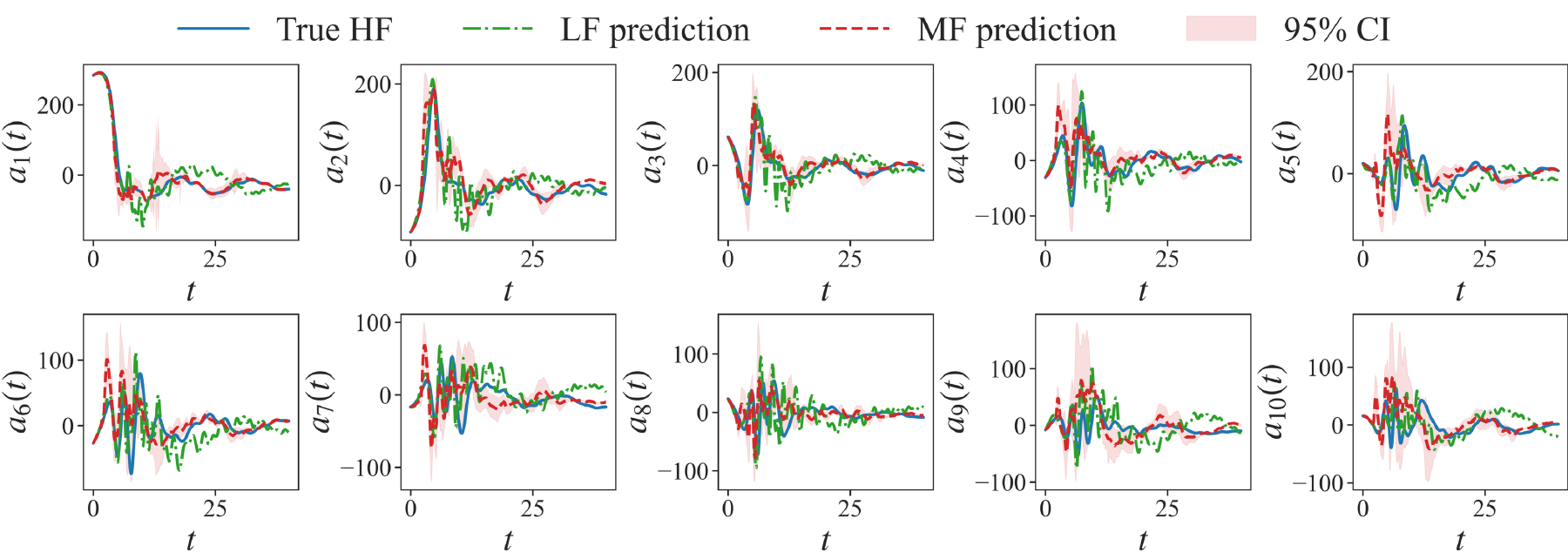} 
  \caption{Probabilistic MF closure correction by the DL method for the double shear layer problem for $\mathrm{Re}=1500$}
  \label{fig:dl_double_shear}
\end{figure}

\Cref{fig:dl_double_shear} presents the DL-based MF prediction results for $\mathrm{Re}=1500$. The figure compares the true HF POD coefficients, the LF ROM predictions, and the probabilistic MF predictions. The shaded region indicates the 95\% credible interval (CI) estimated from the MF ensemble.
Overall, the DL-based MF model captures the dominant temporal evolution of the POD coefficients across the retained modes. Compared with the LF ROM, the MF prediction generally provides improved agreement with the HF solution, particularly at later times, indicating that the DL model can correct the LF representation and recover important features of the HF dynamics.
However, during the early transient stage, where the coefficient trajectories exhibit rapid oscillations and larger amplitudes, some discrepancies between the MF prediction and the HF solution remain, suggesting that directly learning the HF coefficients from LF inputs is still challenging during rapidly varying temporal dynamics.
From the uncertainty quantification perspective, the predictive uncertainty tends to increase during time intervals with larger prediction errors, implying that the DL-based MF ensemble can reflect, to some extent, the varying level of prediction difficulty over time.

\begin{figure}[h]
  \centering
  \includegraphics[width=\textwidth]{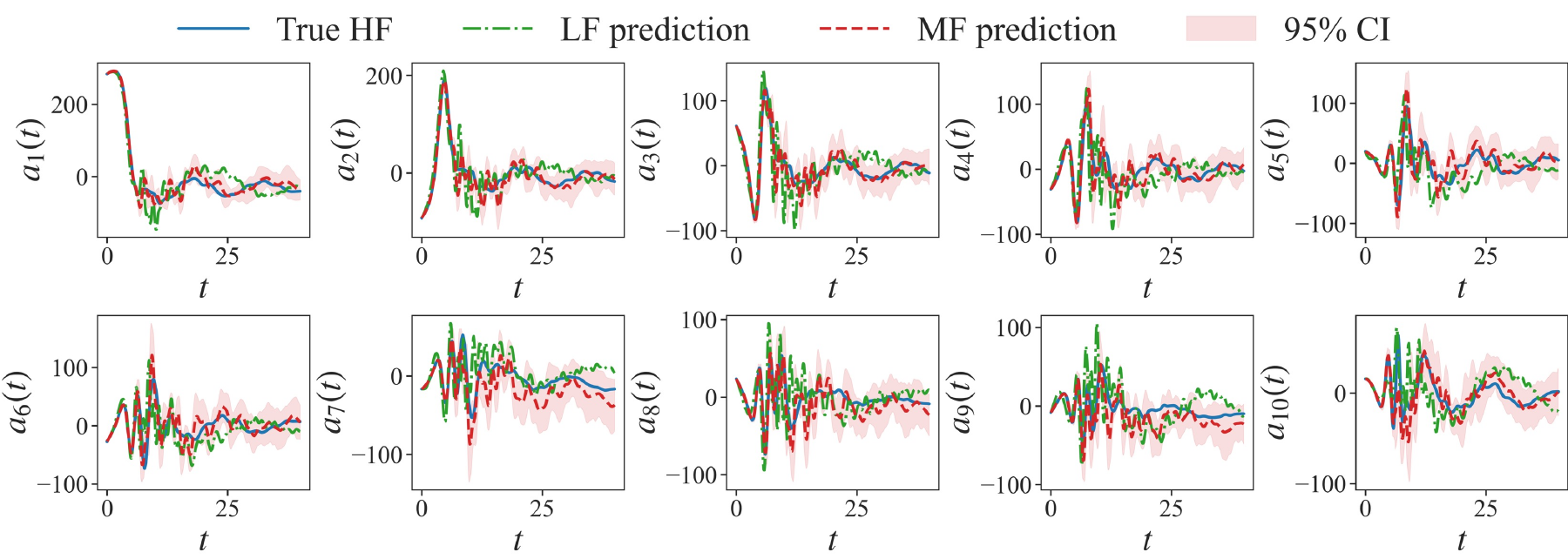} 
  \caption{Probabilistic MF closure correction by the RL method for the double shear layer problem at $Re=1500$}
  \label{fig:rl_double_shear}
\end{figure}

\Cref{fig:rl_double_shear} presents the prediction results obtained using the RL formulation for $\mathrm{Re}=1500$. Compared with the DL results, the RL-based MF prediction shows substantially improved agreement with the true HF trajectories across both low- and high-order POD coefficients. The ensemble-mean MF prediction closely follows the main temporal evolution of the HF solution over the entire time horizon, including the initial transient stage and the later-time oscillatory response. This improvement is particularly evident for modes where the LF ROM exhibits noticeable amplitude and phase discrepancies.
The superior performance of the RL formulation can be attributed to its residual-learning structure. Instead of directly learning the full HF POD coefficients, the RL model learns the discrepancy between the LF and HF states. The LF--HF residual is generally lower in magnitude than the full HF state. As a result, the CNF only needs to model the correction required to compensate for the LF model bias, rather than reconstructing the complete HF ROM state from the LF input, making the learning task more stable and allowing the MF model to better preserve the physically meaningful temporal evolution provided by the LF ROM while correcting its systematic errors.

We quantitatively evaluate the DL and RL strategies using the relative $\ell_2$ error (RL2E) and log-predictive probability (LPP) metrics.
Both component-wise and overall metrics are reported.
Let $a_{n,k}$ denote the ground-truth HF POD coefficient at time step $n=1,\ldots,N_t$ and POD mode $k=1,\ldots,R$. Furthermore, let $\mu_{n,k}$ and $\sigma_{n,k}^{2}$ denote the ensemble mean and variance, respectively, of the corresponding MF predictions computed using \Cref{eq:dl_mean,eq:dl_std_component}.
For the $k$-th POD mode, the component-wise and overall RL2E are defined as
\begin{gather}
  \mathrm{RL2E}(k)
  =\sqrt{\sum\nolimits_{n=1}^{N_t}\left(\mu_{n,k}-a_{n,k}\right)^2}
  \bigg/
  \sqrt{\sum\nolimits_{n=1}^{N_t}a_{n,k}^{2}},\\
  \mathrm{RL2E}_{\mathrm{overall}}
  =\sqrt{\sum\nolimits_{k=1}^{R}\sum\nolimits_{n=1}^{N_t}
    \left(\mu_{n,k}-a_{n,k}\right)^2}
  \bigg/\sqrt{\sum\nolimits_{k=1}^{R}\sum\nolimits_{n=1}^{N_t}a_{n,k}^{2}}.
\end{gather}
The component-wise metric $\mathrm{RL2E}(k)$ measures the deterministic prediction error for an individual POD coefficient, whereas $\mathrm{RL2E}_{\mathrm{overall}}$ measures the discrepancy across the complete reduced-order trajectory. Both metrics are dimensionless, and smaller values indicate better agreement between the predictive mean and the ground-truth HF solution.

Assuming a Gaussian predictive distribution for each POD coefficient, the component-wise and overall LPP are defined as
\begin{gather}
  \mathrm{LPP}(k)=-\frac{1}{2N_t}\sum\nolimits_{n=1}^{N_t}
  \left[\log\left(2\pi\sigma_{n,k}^{2}\right)
    +\frac{\left(a_{n,k}-\mu_{n,k}\right)^2}{
      \sigma_{n,k}^{2}}
  \right],\\
  \mathrm{LPP}_{\mathrm{overall}}
  =\frac{1}{R}\sum\nolimits_{k=1}^{R}\mathrm{LPP}(k).
\end{gather}
Unlike the RL2E, the LPP evaluates both the accuracy of the predictive mean and the quality of the predicted uncertainty. Larger LPP values indicates a better probabilistic prediction with a more favorable balance between accuracy and uncertainty calibration.

\begin{figure}[H]
  \centering
  \begin{subfigure}[b]{0.47\textwidth}
    \centering
    \includegraphics[width=\textwidth]{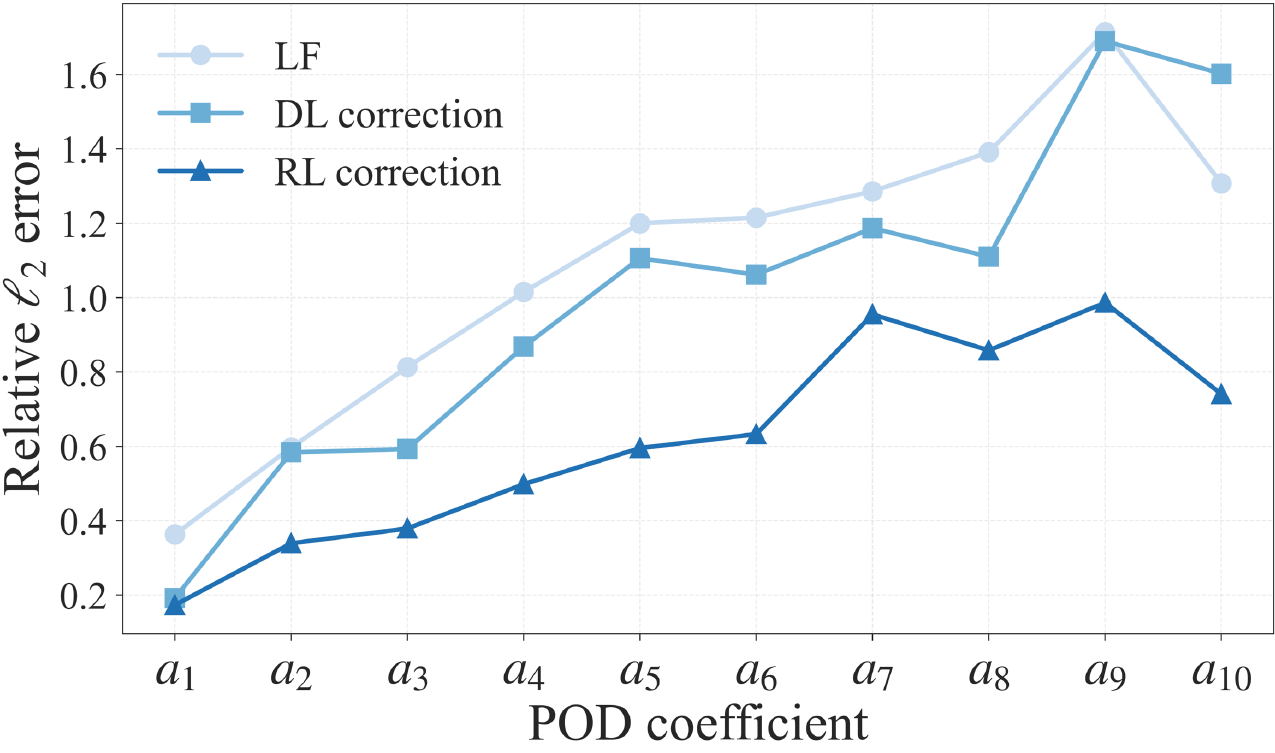}
    \caption{\(\text{RL2E}(k)\).}
    \label{fig:rl_double_shear_l2}
  \end{subfigure}
  \hfill
  \begin{subfigure}[b]{0.475\textwidth}
    \centering
    \includegraphics[width=\textwidth]{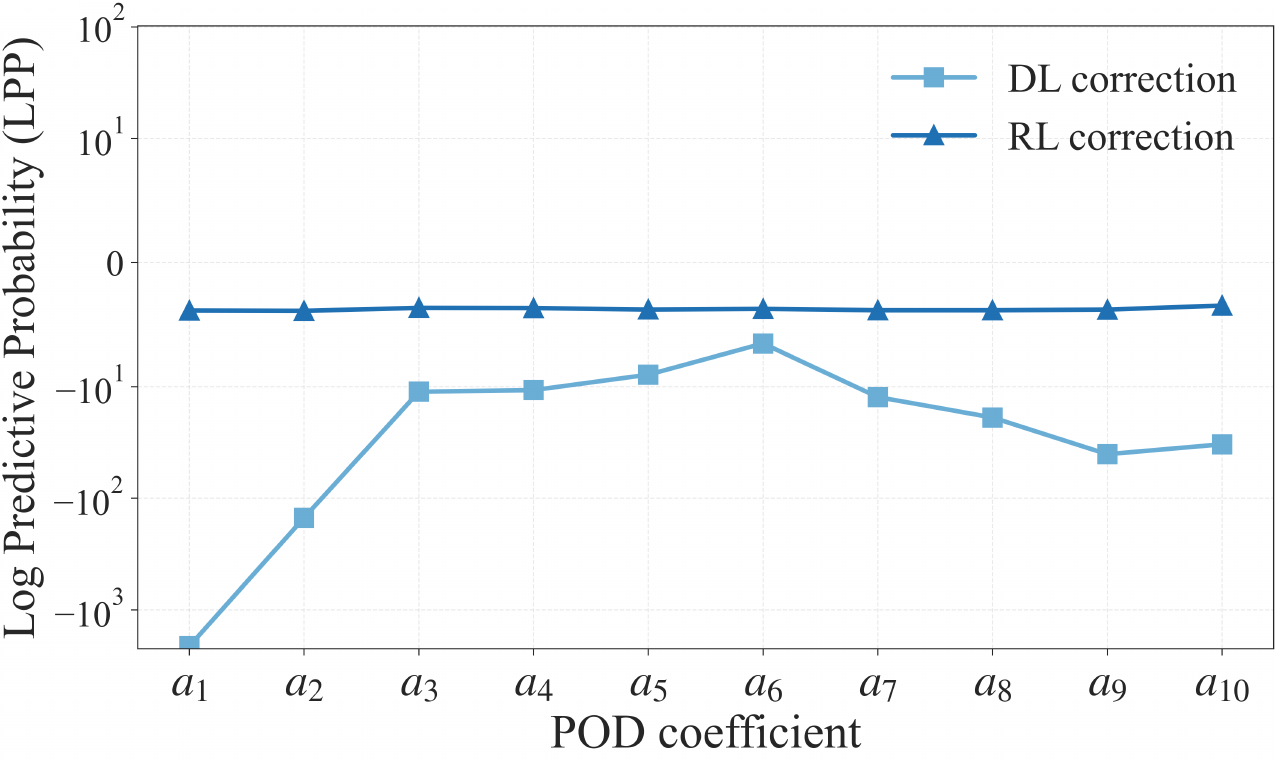}
    \caption{\(\text{LPP}(k)\).}
    \label{fig:rl_double_shear_lpp}
  \end{subfigure}

  \caption{Component-wise performance of the DL and RL corrections for the double shear layer problem at $\mathrm{Re}=1500$.}
  \label{fig:rl_double_shear_metrics}
\end{figure}

\Cref{fig:rl_double_shear_l2} and \Cref{tab:double_l2_error} compare the coefficient-wise RL2E of the LF ROM, DL correction, and RL correction for the POD coefficients $a_1$--$a_{10}$. Overall, the relative error tends to increase from lower- to higher-order POD coefficients for all three methods because higher-order POD modes usually contain more complex and rapidly varying dynamical features, making them more difficult to predict accurately.

The RL correction strategy consistently yields the lowest RL2E for all POD coefficients, indicating superior predictive accuracy. The overall RL2E is reduced from 0.691 for the LF ROM to 0.376 for the RL correction, corresponding to an improvement of approximately $45.6\%$, demonstrating that the RL formulation can effectively correct the systematic discrepancy between the LF ROM and HF dynamics.
The DL correction also improves over the LF ROM for most coefficients, reducing the overall error from 0.691 to 0.600, which corresponds to an improvement of approximately $13.2\%$. However, its accuracy remains noticeably lower than that of the RL correction. In particular, for $a_{10}$, the DL correction produces a larger error than the LF ROM.

\begin{table}[htbp]
\centering

\begin{subtable}[t]{\textwidth}
    \centering
    \caption{\( \text{RL2E}(k) \) and overall for LF ROM, DL correction, and RL correction.}
    \label{tab:double_l2_error}
    \resizebox{\linewidth}{!}{%
    \begin{tabular}{lccccccccccc}
        \toprule
        \textbf{Method}
        & \(a_1\) & \(a_2\) & \(a_3\) & \(a_4\) & \(a_5\)
        & \(a_6\) & \(a_7\) & \(a_8\) & \(a_9\) & \(a_{10}\)
        & Average \\
        \midrule
        LF
        & 0.363 & 0.597 & 0.814 & 1.015 & 1.200
        & 1.215 & 1.286 & 1.391 & 1.714 & 1.308
        & 0.691 \\
        DL
        & 0.192 & 0.584 & 0.593 & 0.868 & 1.106
        & 1.062 & 1.187 & 1.110 & 1.690 & 1.603
        & 0.600 \\
        RL
        & 0.176 & 0.348 & 0.393 & 0.512 & 0.605
        & 0.643 & 0.991 & 0.880 & 1.017 & 0.740
        & 0.376 \\
        \bottomrule
    \end{tabular}%
    }
\end{subtable}
\vspace{0.8em}
\begin{subtable}[t]{\textwidth}
    \centering
    \caption{\( \text{LPP}(k) \) and overall LPP values of DL and RL corrections.}
    \label{tab:double_lpp}
    \resizebox{\linewidth}{!}{%
    \begin{tabular}{lccccccccccc}
        \toprule
        \textbf{Method}
        & \(a_1\) & \(a_2\) & \(a_3\) & \(a_4\) & \(a_5\)
        & \(a_6\) & \(a_7\) & \(a_8\) & \(a_9\) & \(a_{10}\)
        & Average \\
        \midrule
        DL
        & -2110.69 & -149.22 & -11.12 & -10.74 & -9.03
        & -6.52 & -12.41 & -18.90 & -40.30 & -32.83
        & -240.18 \\
        RL
        & -3.96 & -3.92 & -3.67 & -3.68 & -3.77
        & -3.74 & -3.85 & -3.85 & -3.83 & -3.47
        & -3.78 \\
        \bottomrule
    \end{tabular}%
    }
\end{subtable}

\caption{Component-wise and overall performance of the LF ROM, DL correction, and RL correction for the double shear layer problem at $\mathrm{Re}=1500$.}
\label{tab:double_shear_metrics}
\end{table}

\Cref{fig:rl_double_shear_lpp} and \Cref{tab:double_lpp} compare the coefficient-wise LPP values of the DL and RL correction methods for the POD coefficients \(a_1\)--\(a_{10}\).
Since a larger LPP value indicates a better probabilistic prediction, the RL correction clearly outperforms the DL correction for all POD modes.
The LPP values of the RL correction remain stable, ranging approximately from \(-4.0\) to \(-3.5\), with an overall value of \(-3.7752\). 
In contrast, the DL correction yields substantially smaller LPP values for all coefficients, with an overall value of \(-240.176\).
The degradation is especially pronounced for the leading modes, such as \(a_1\).
The superior performance of the RL correction can be attributed to its residual-learning formulation.
By learning the LF--HF discrepancy rather than directly mapping the LF coefficients to the HF coefficients, the RL correction preserves the dominant LF dynamical structure while correcting the remaining model-form error.

\subsection{A vortex merger problem}\label{sec:vortex}

A two-dimensional vortex merger problem is considered as the second benchmark to further assess the proposed MF framework.
This problem describes the evolution of two initially co-rotating vortices in close proximity that gradually interact and merge into a larger coherent vortex structure \cite{meunier2005physics,pawar2021model}.
It captures fundamental vortex dynamics that arise in a wide range of applications, including astrophysics, meteorology, and geophysical flows.

The same vorticity--streamfunction formulation introduced in \Cref{sec:double_shear} is used to describe the flow evolution.
The computational domain is $(x,y)\in[0,2\pi]\times[0,2\pi]$ with periodic boundary conditions.
The initial vorticity field is prescribed as two Gaussian-distributed vortices,
\begin{equation}
\begin{aligned}
\omega(x,y,0)
={}&\exp\left\{-\rho\left[(x-x_1)^2+(y-y_1)^2\right]
\right\}
\\&+\exp\left\{-\rho\left[(x-x_2)^2+(y-y_2)^2\right]
\right\}.
\end{aligned}
\end{equation}
where $\rho$ is the interaction constant.
In this study, we choose $\rho=\pi$, and we set the two initial vortex centers at coordinates
$(x_1,y_1)=(3\pi/4,\pi)$ and $(x_2,y_2)=(5\pi/4,\pi)$.
Following the procedure described in the double shear layer example, the HF POD coefficients are obtained by projecting the full-order vorticity snapshots onto the global POD basis, while the LF coefficient trajectories are generated from the GPOD model.

\subsubsection{CNF model training}\label{sec:CNF_train}
Similarly to \Cref{sec:double_CNF_train}, the ROM is constructed by retaining 10 time-dependent POD coefficients, corresponding to an RIC of \(\sim 98 \%\), and data is generated by simulating over 401 time steps for 10 Reynolds number values.
RIC as a function of the number of retained modes is shown in \Cref{fig:RIC_vortex_merger}.
Therefore, both the LF and HF datasets are represented as tensors of size $10 \times 401 \times 10$, where the three dimensions denote the Reynolds number, temporal snapshot, and POD mode, respectively. Because the proposed framework performs recursive prediction for both the DL and RL formulations as described by \Cref{eq:dl_recursive_sequence,eq:rl_recursive_sequence}, the data is reshaped into matrices of size $4010 \times 10$.

As in the double shear layer example, only snapshots over the time interval \(t \in [0,38] \) are used to construct the paired data for CNF training, and terminal interval \(38 < t \le 40\) is excluded from training.
The trained model is subsequently rolled out over the complete interval \(t \in [0,40]\), such that the final two time units represent a temporal extrapolation regime.
We employ the same CNF architecture as that described in \Cref{sec:double_CNF_train}.
Specifically, we employ a MAF architecture consisting of five MADE-based autoregressive blocks with output dimension of 10 and two hidden layers with 64 neurons each. The model is trained using the Adam optimizer over 20,000 iterations with a learning rate of $10^{-4}$ and a batch size of 128.

\subsubsection{Results of closure correction}\label{sec:results}

After training, the CNF model is used to perform probabilistic sequential prediction over the full time horizon of 401 snapshots.
For each Reynolds number, the prediction is initialized using the corresponding HF POD coefficients at the first snapshot, and the subsequent states are generated recursively in time.
To quantify predictive uncertainty, this sequential rollout is repeated 200 times with different random seeds, resulting in 200 ensemble trajectories.
It should be emphasized that only one sample is drawn from the CNF at each time step within a given trajectory; the ensemble is constructed by repeating the entire recursive prediction procedure 200 times rather than drawing multiple samples simultaneously at a single time step, yielding an ensemble-based approximation of the probabilistic evolution of the POD coefficients.

\begin{figure}[h]
  \centering
  \includegraphics[width=\textwidth]{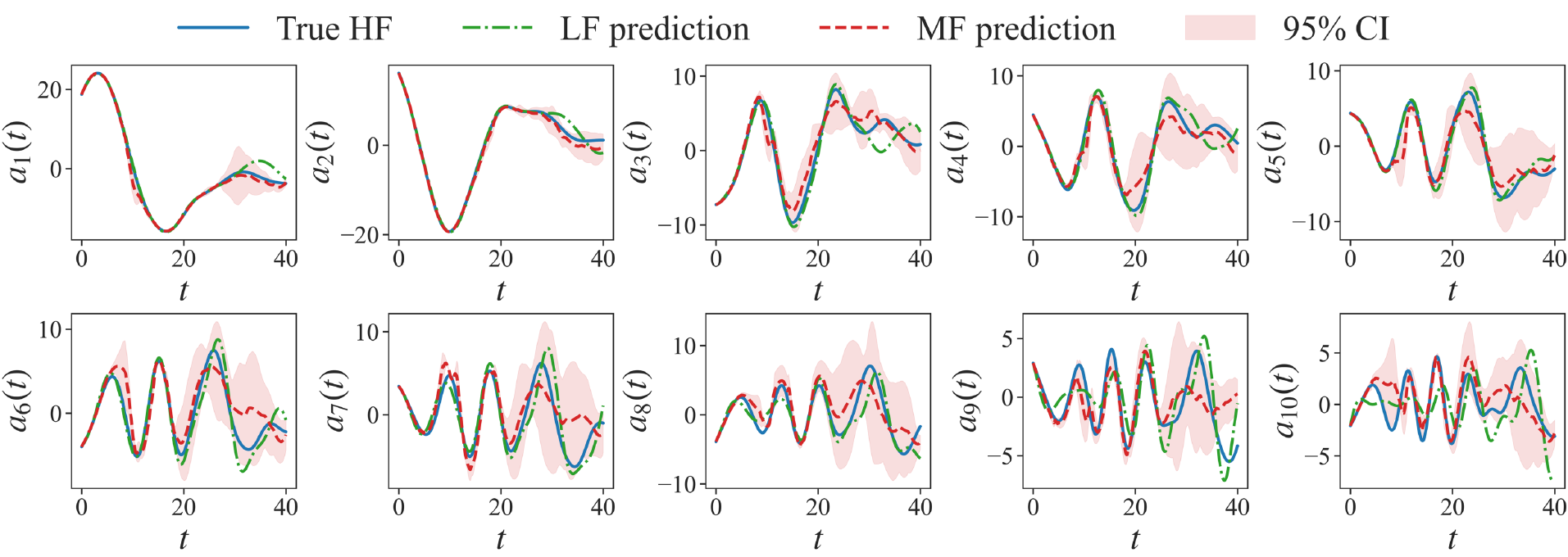} 
  \caption{Probabilistic MF closure correction by the DL method for the vortex merger problem at $\mathrm{Re}=5000$}
  \label{fig:DL}
\end{figure}

\Cref{fig:DL} shows the MF predictions obtained by the DL method for the Reynolds number $\mathrm{Re}=5000$.
In each subplot, the blue solid line denotes the true HF POD coefficient, the green dash-dot line denotes the LF prediction, the red dashed line denotes the mean MF prediction, and the shaded region represents the 95\% CI of the MF prediction.
It can be observed that the mean MF prediction agrees with the true HF response better than the LF prediction across most POD coefficients, demonstrating the effectiveness of the DL-based closure correction. The improvement is particularly evident in the later stage for $a_1$ and $a_2$, where the MF prediction remains much better aligned with the true HF trajectory. For the higher-order POD coefficients, which display stronger oscillations and more irregular fluctuations, the DL method is still able to capture the main complex dynamical patterns. This improved agreement is especially clear for $a_9$ and $a_{10}$ during the first 20 seconds, where the MF prediction follows the true HF trajectory more closely than the LF prediction.
Regarding uncertainty quantification, the predicted uncertainty generally increases as time advances, reflecting the accumulation of error during recursive sequence prediction.
Larger uncertainty bands also tend to appear in regions with larger prediction discrepancies, such as during $t=30$--$40$~s for $a_6$--$a_8$.
Nevertheless, most snapshots of the true HF response remain enclosed within the predicted uncertainty bands.

\begin{figure}[H]
  \centering
  \includegraphics[width=\textwidth]{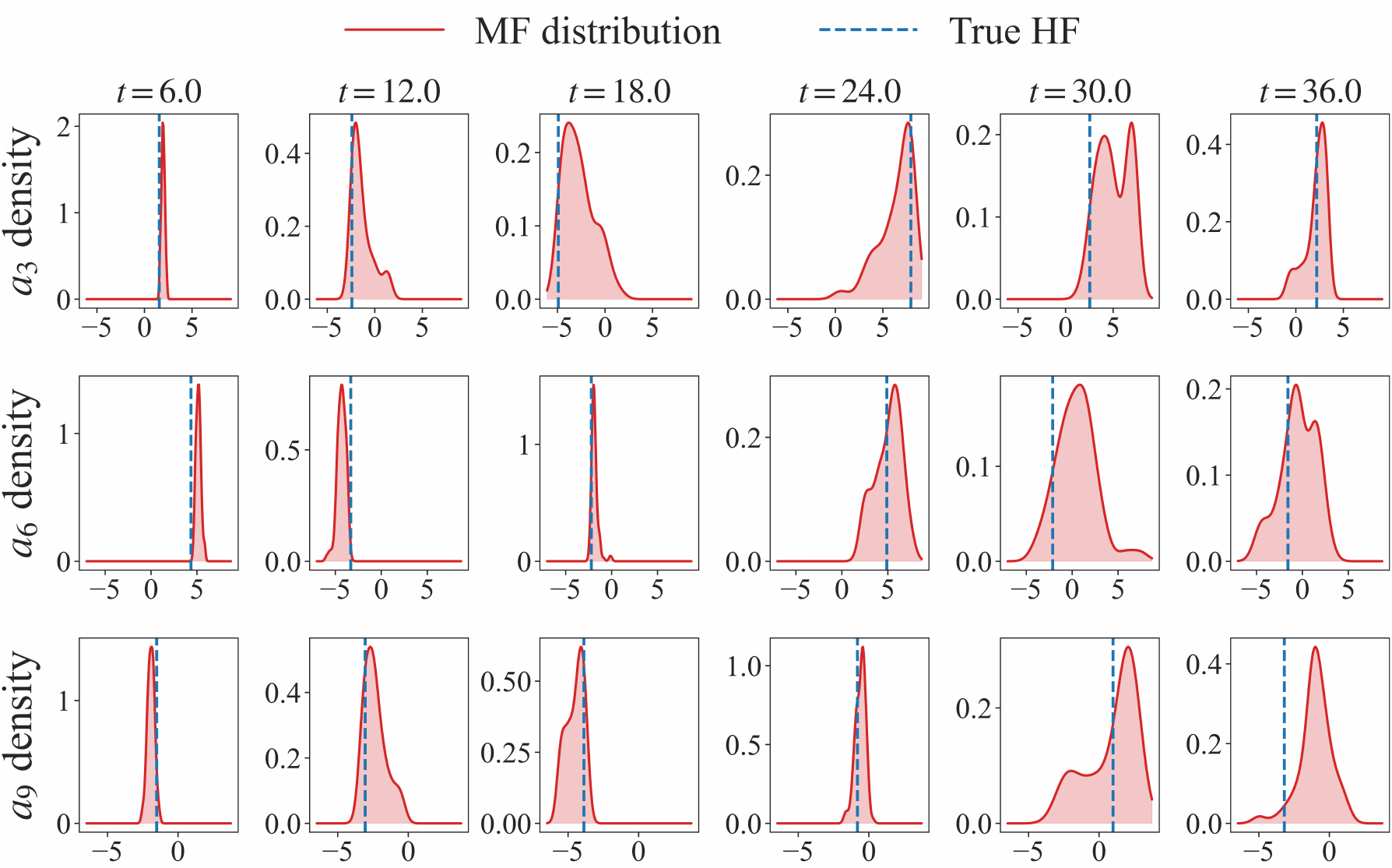} 
  \caption{Predicted MF probability distributions for the POD coefficients $a_3$, $a_6$, and $a_9$ at selected times computed via the DL method for the vortex merger problem at \(\mathrm{Re}=5000\).}
  \label{fig:DL_dist}
\end{figure}

\Cref{fig:DL_dist} illustrates the evolution of the predicted probability distributions of the MF solution at selected times for the POD coefficients $a_3$, $a_6$, and $a_9$. The figure shows how the uncertainty distribution changes as the recursive prediction progresses from $t=6$~s to $t=36$~s.

Overall, the distributions tend to become broader at later times, indicating an increase in predictive uncertainty as the sequence evolves.
This trend is consistent with the recursive nature of the forecasting process, in which approximation errors and stochastic correction errors accumulate over time.
Furthermore, the predicted distributions are not always Gaussian.
At later times, several coefficients exhibit clearly skewed or multi-modal distributions, particularly around $t=24$--36~s, suggesting that the system may admit multiple plausible future states under the learned probabilistic correction.
The proposed generative framework is able to represent these complex distributional patterns.

\begin{figure}[H]
  \centering
  \includegraphics[width=\textwidth]{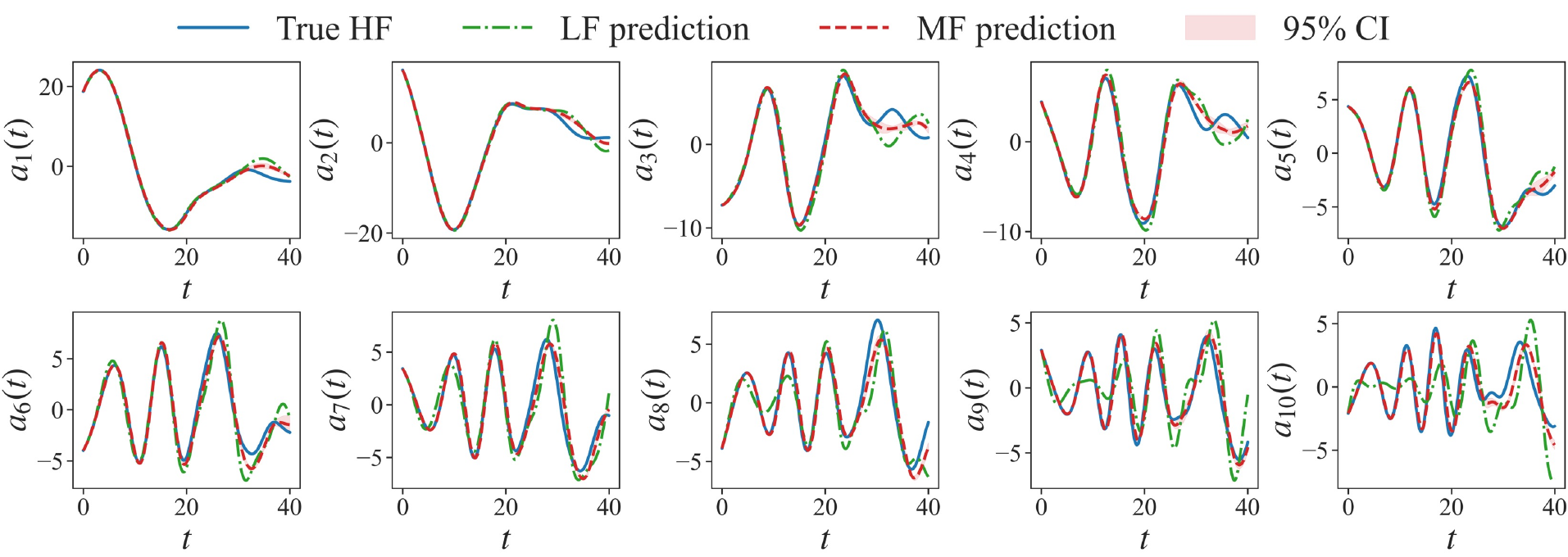} 
  \caption{Probabilistic MF closure correction by the RL method for the vortex merger problem at $\mathrm{Re}=5000$}
  \label{fig:RL}
\end{figure}

\Cref{fig:RL} presents the MF predictions obtained by the RL method for $\mathrm{Re}=5000$.
It is evident that the RL method substantially improves prediction accuracy relative to the LF model.
The MF mean prediction remains closely aligned with the true HF response for most POD coefficients throughout the time horizon.
This improvement is observed not only for the low-order modes, which dominate the global response, but also for the higher-order modes, which exhibit stronger oscillations and more complex temporal patterns.
In particular, the RL method shows strong correction capability for the higher POD coefficients, where the LF prediction tends to deviate more noticeably from the true HF dynamics.
These results indicate that the RL formulation is effective in learning the discrepancy between LF and HF trajectories and in recovering the dynamics not captured by the ROM.

\begin{figure}[ht]
  \centering
  \includegraphics[width=\textwidth]{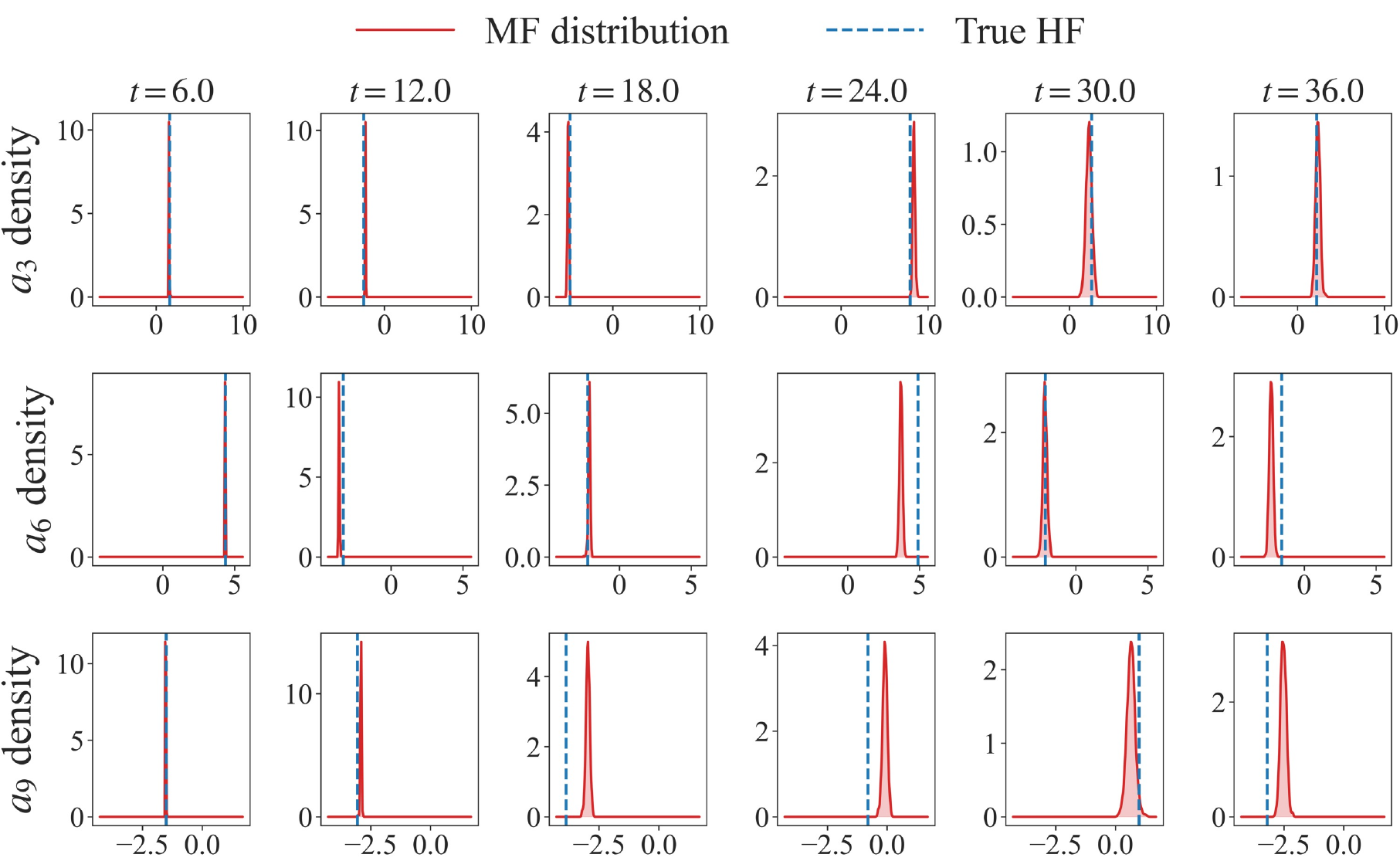} 
  \caption{Predicted MF probability distributions for the POD coefficients $a_3$, $a_6$, and $a_9$ at selected times computed via the RL method for the vortex merger problem at \(\mathrm{Re}=5000\).}
  \label{fig:RL_dist}
\end{figure}

Compared with the DL results in \Cref{fig:DL}, the uncertainty bands produced by the RL method are noticeably narrower and are almost visually indistinguishable.
While the MF mean predictions are overall accurate, the tight 95\% CI indicate an overconfident uncertainty estimates.
\Cref{fig:RL_dist} shows the evolution of the predicted probability distributions of the MF solution obtained by the RL method at selected times for the POD coefficients $a_3$, $a_6$, and $a_9$.
It can alos be seen here that the predicted uncertainty remains very small throughout the simulation.
For these three POD coefficients, the distributions are highly concentrated around their mean values, indicating that the RL method yields very confident predictions, which is consistent with the narrow CI observed in the time-history results.
Nevertheless, as time moves forward, the distributions still exhibit a gradual increase in spread, suggesting that uncertainty accumulation is still present in the recursive forecasting process, although its magnitude is much smaller than that in the DL case.

In addition, the RL-predicted distributions remain generally unimodal and close to Gaussian at all selected times, without pronounced multi-modal or skewed distributions seen in the DL case, suggesting that the residual-based formulation leads to a simpler uncertainty structure likely because the model learns a correction term that is smaller in magnitude and less variable than the full HF state.

\subsubsection{Effect of noise on MF predictions}\label{sec:noise}

Even though the mean RL prediction is accurate, its overconfident uncertainty estimates are undesirable because they don't cover the true HF snapshots over the prediction window.
One possible explanation is that, by construction, the training of the CNF only considers one step of rollout (from \(t_{k_t - 1}\) to \(t_{k_t}\) for the \(k\)th data pair, cf. \Cref{sec:RL}) and thus doesn't account for the accumulation of error over multiple rollout steps.
To address this limitation, this section considers the effect of introducing different noise levels into the LF input during training. The motivation is to examine whether perturbing the LF input can partially account for
uncertainties due to rollout error accumulation
and thus improve uncertainty quantification in the RL framework. Specifically, additive Gaussian noise with standard deviation levels of 5\%, 10\%, 15\%, and 20\% is applied to the LF input.

\begin{figure}[http]
  \centering
  \begin{subfigure}[b]{\textwidth}
    \centering
    \includegraphics[width=\textwidth]{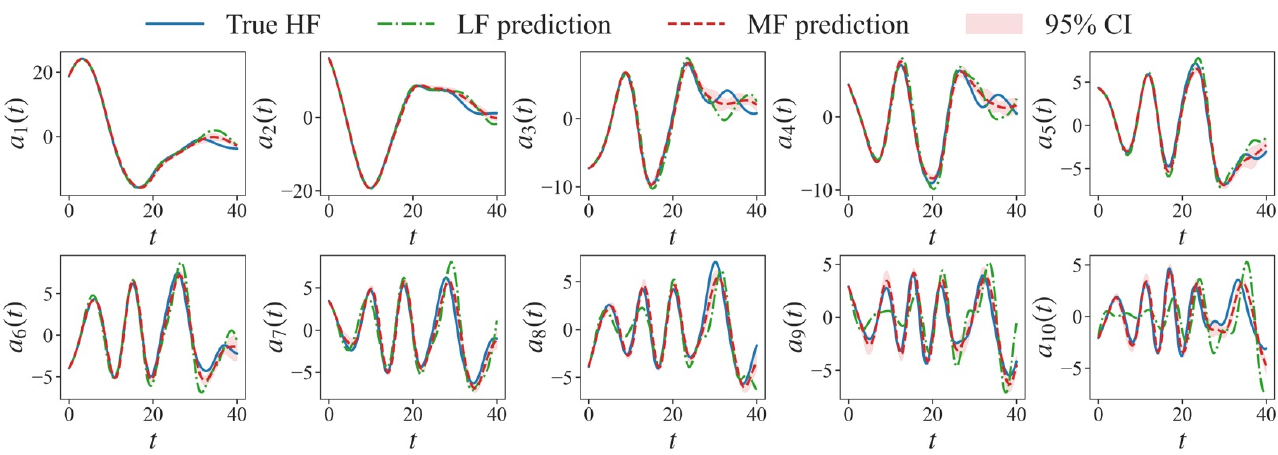}
    \caption{Noise level: 5\%.}
    \label{fig:RL_0.05}
  \end{subfigure}
  
  \vspace{0.5cm}
  
  \begin{subfigure}[b]{\textwidth}
    \centering
    \includegraphics[width=\textwidth]{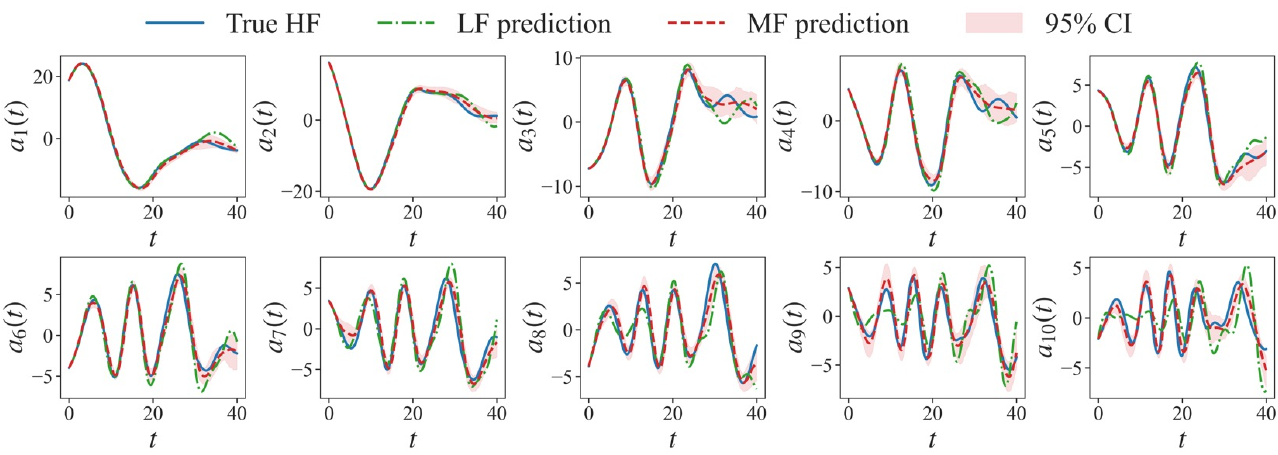}
    \caption{Noise level: 10\%.}
    \label{fig:RL_0.10}
  \end{subfigure}
  \caption{MF predictions obtained by the RL method for the vortex merger problem at \(\mathrm{Re}=5000\) with 5\% and 10\% Gaussian noise added to the LF input}
  \label{fig:RL_noise_1}
\end{figure}
\begin{figure}[http]
  \centering
  \begin{subfigure}[b]{\textwidth}
    \centering
    \includegraphics[width=\textwidth]{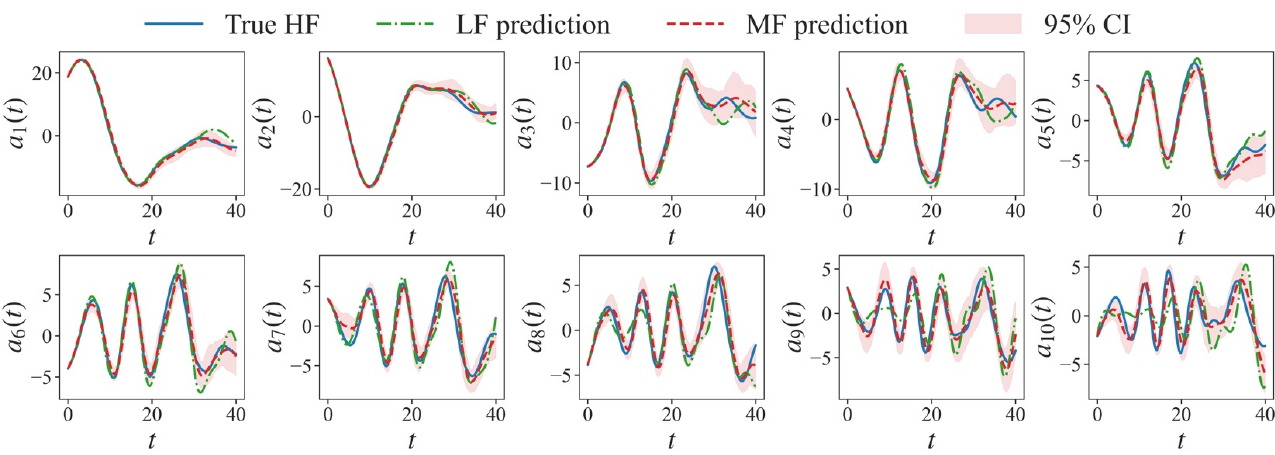}
    \caption{Noise level: 15\%.}
    \label{fig:RL_0.15}
  \end{subfigure}
  
  \vspace{0.5cm}
  
  \begin{subfigure}[b]{\textwidth}
    \centering
    \includegraphics[width=\textwidth]{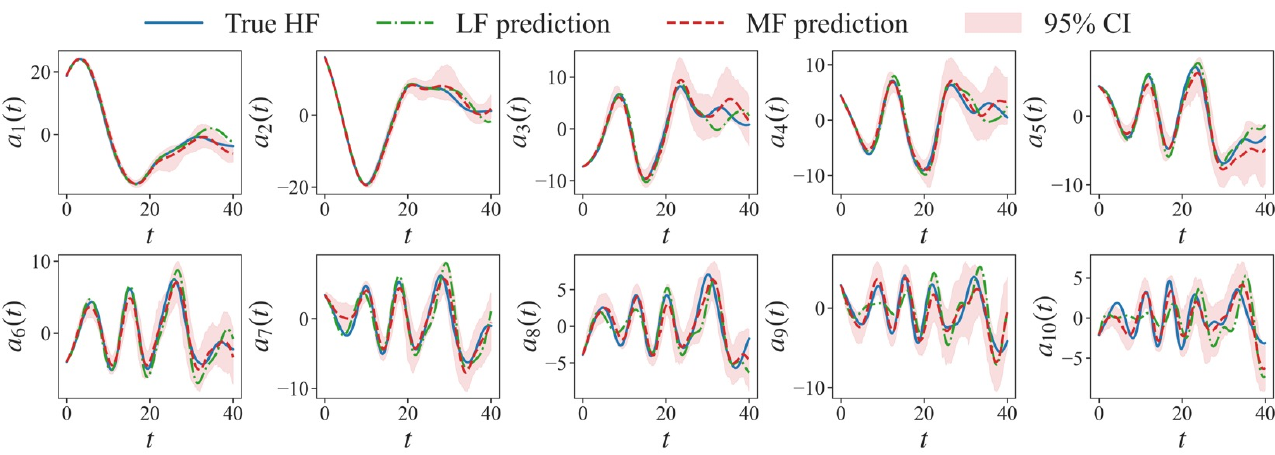}
    \caption{Noise level: 20\%.}
    \label{fig:RL_0.20}
  \end{subfigure}
  \caption{MF predictions obtained by the RL method for the vortex merger problem at \(\mathrm{Re}=5000\) with 15\% and 20\% Gaussian noise added to the LF input.}
  \label{fig:RL_noise_2}
\end{figure}

\cref{fig:RL_noise_1,fig:RL_noise_2} shows the results obtained by the RL method for different noise levels.
Overall, these results show that introducing Gaussian noise into the LF input increases the predicted uncertainty in the RL framework and improves the coverage of the true HF trajectories.
However, this improvement in uncertainty coverage comes at the cost of reduced predictive accuracy, as the MF mean prediction gradually deviates more from the true HF solution with increasing noise level, especially for the higher-order POD coefficients that exhibit more complex and oscillatory behavior.

Taken together, these results reveal a clear trade-off between accuracy and uncertainty quantification in the RL framework with noisy LF inputs. A small noise level such as 5\% is insufficient to fully address overconfidence, whereas large noise levels such as 20\% overly broaden the uncertainty bands while noticeably reduce the accuracy of the MF mean prediction. Intermediate noise levels, particularly around 10\% to 15\%, appear to offer a more reasonable balance between maintaining accurate mean predictions and achieving more reliable uncertainty coverage.

In addition to the RL2E and LPP metrics, we also compute the empirical coverage of the nominal CIs to directly assess uncertainty calibration.
Ideally, a well-calibrated probabilistic model should produce a coverage close to the nominal level of 0.95. A coverage substantially lower than 0.95 indicates underestimation of uncertainty, i.e., overconfidence, whereas a coverage much higher than 0.95 suggests that the predicted uncertainty bands are overly conservative.
We consider a set of nominal coverage levels \(\mathcal{P}=\{0, 0.1,0.2,\ldots,0.9,0.95, 1\}\).
For each nominal level \(p\in\mathcal{P}\), the corresponding prediction interval is defined as
\begin{equation*}
\mathcal{C}_{n,k}(p)=\left[q_{n,k}^{(1-p)/2},\,q_{n,k}^{(1+p)/2}\right],
\end{equation*}
where \(q_{n,k}^{(1-p)/2}\) and \(q_{n,k}^{(1+p)/2}\) denote the lower and upper empirical quantiles computed from the predictive samples of the \(k\)th POD coefficient at time step \(t_n\), respectively.
An indicator function is then introduced as \(
I_{n,k}(p)=\mathbf{1}\left(
a_{n,k}\in \mathcal{C}_{n,k}(p)
\right)\), where \(I_{n,k}(p)=1\) if the true HF value falls inside the predicted interval at nominal level \(p\), and \(I_{n,k}(p)=0\) otherwise.
The empirical coverage corresponding to the nominal level \(p\) is computed by averaging this indicator over all time steps and POD coefficients:
\begin{equation*}
\widehat{p}_{\mathrm{emp}}(p)
=\frac{1}{N_t R}
\sum_{n=1}^{N_t}
\sum_{k=1}^{R}
I_{n,k}(p).
\end{equation*}
A calibration curve can then be computed by plotting the nominal coverage \(p\) against the empirical coverage \(\widehat{p}_{\mathrm{emp}}(p)\). Ideally, a well-calibrated probabilistic model should produce a calibration curve close to the diagonal line \(\widehat{p}_{\mathrm{emp}}(p)=p\). If the curve lies below the diagonal, the empirical coverage is lower than the nominal coverage, indicating that the predicted uncertainty intervals are too narrow and the model is overconfident. In contrast, if the curve lies above the diagonal, the empirical coverage is higher than the nominal coverage, suggesting that the predicted uncertainty intervals are overly conservative. Therefore, the calibration curve provides a more comprehensive assessment of uncertainty reliability than a single 95\% coverage value, since it evaluates calibration behavior over multiple confidence levels.

\begin{figure}[http]
  \centering
  \begin{subfigure}[b]{0.475\textwidth}
    \centering
    \includegraphics[width=\textwidth]{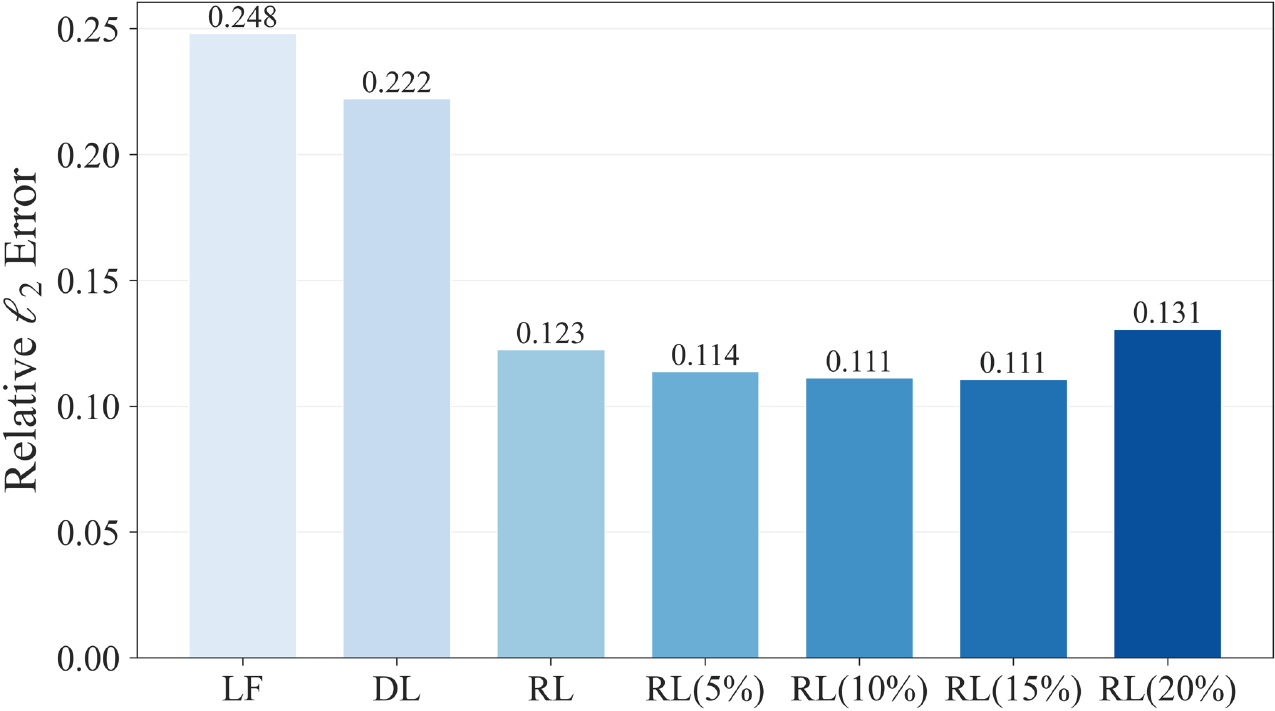}
    \caption{Overall \(\text{RL2E}\).}
    \label{fig:l2}
  \end{subfigure}
  \hfill
  \begin{subfigure}[b]{0.475\textwidth}
    \centering
    \includegraphics[width=\textwidth]{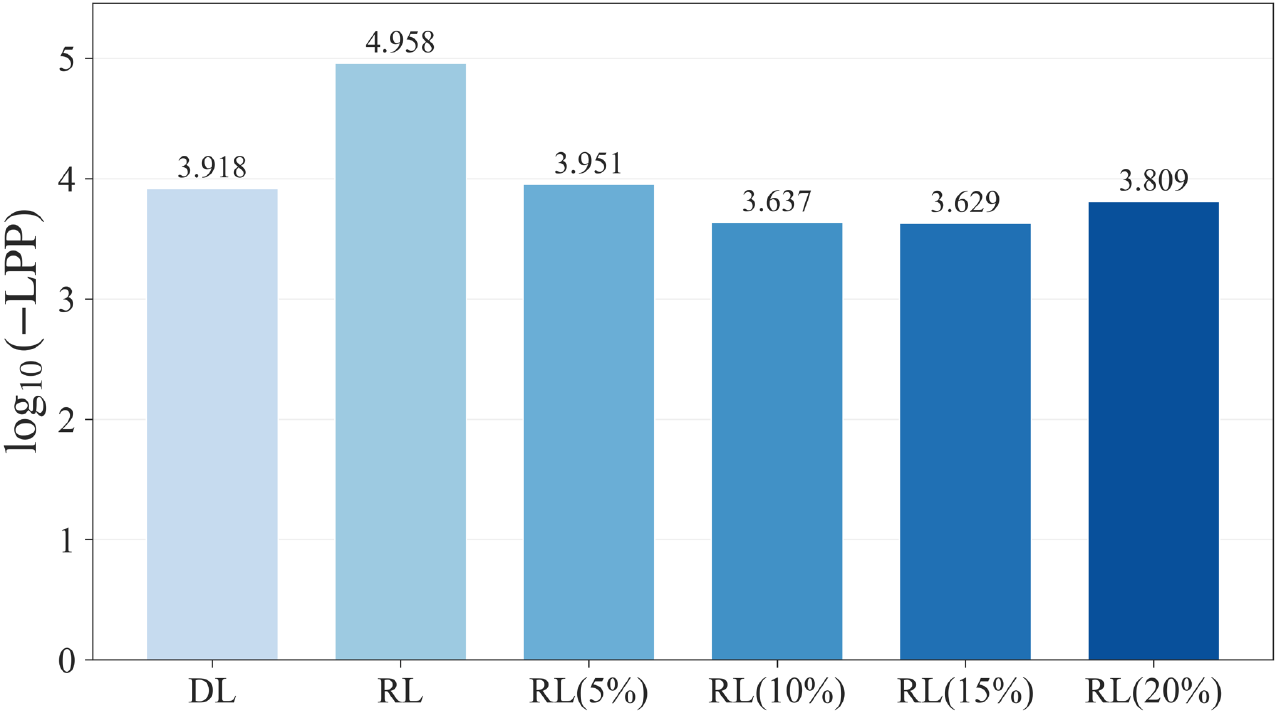}
    \caption{Overall \( \log_{10}(-\text{LPP}) \).}
    \label{fig:lpp}
  \end{subfigure}
  \caption{Comparison of the DL and RL methods for the vortex merger problem at $\mathrm{Re}=5000$: (a) Overall \(\text{RL2E}\). (b) Overall \(\log_{10} (\text{-LPP})\). For both panels, smaller values indicate better predictive performance.}
  \label{fig:vortex_merger_metrics}
\end{figure}

\begin{figure}[http]
  \centering
  \begin{subfigure}[b]{0.5\textwidth}
    \centering
    \includegraphics[width=\textwidth]{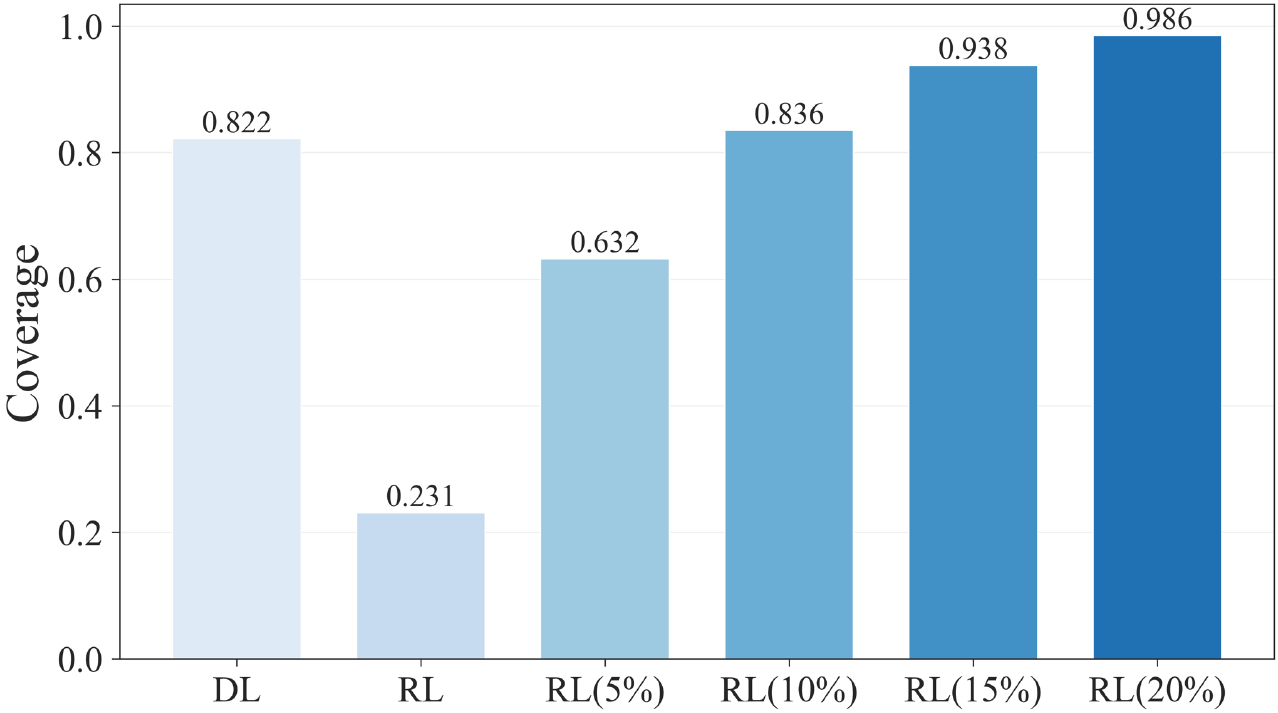}
    \vskip1ex
    \caption{}
    \label{fig:coverage}
  \end{subfigure}
  \begin{subfigure}[b]{0.42\textwidth}
    \centering
    \includegraphics[width=\textwidth]{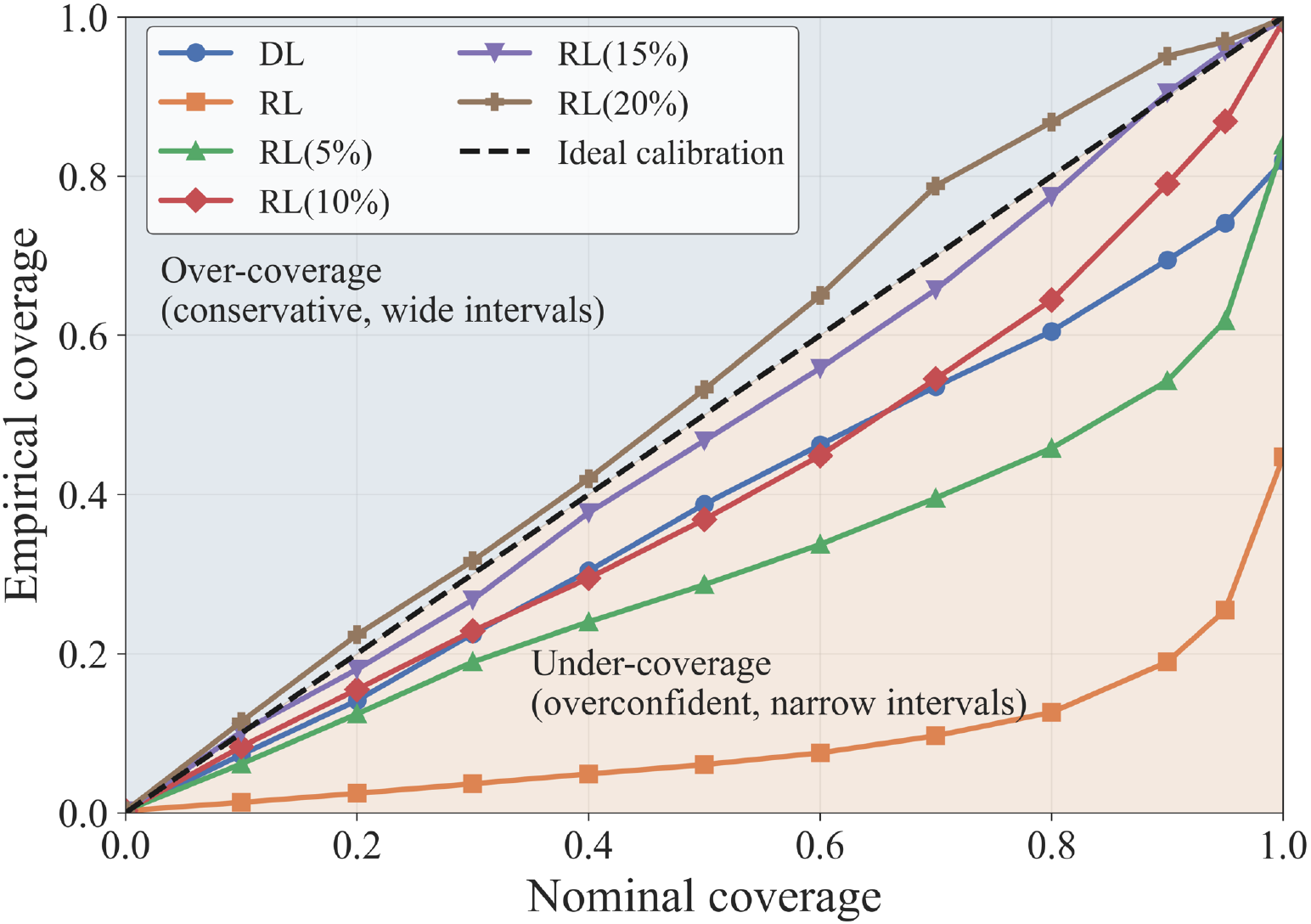}
    \caption{}
    \label{fig:calibration_curve}
  \end{subfigure}

  \caption{Uncertainty calibration of the DL and RL methods for the vortex merger problem at $\mathrm{Re}=5000$: (a) 95\% CI empirical coverage for the DL method and the RL method under different noise levels. (b) Calibration curves for the same methods. Curves below the diagonal indicate overconfident uncertainty estimates, and curves above the diagonal indicate conservative uncertainty estimates.}
  \label{fig:vortex_merger_calibration} 
\end{figure}

\Cref{fig:l2,fig:lpp,fig:coverage} compare the quantitative performance of the LF, DL, and RL methods under different noise levels using the RL2E, LPP, and \(\hat{p}_{\mathrm{emp}}(0.95)\) metrics.
For the RL2E results, the DL method shows only a modest improvement over the LF prediction when the error is aggregated over all POD modes and all time steps.
This is because the overall RL2E may not fully reflect the local improvements achieved by the DL method in several higher-order POD coefficients with more complex dynamics, as observed in \Cref{fig:DL}.
In contrast, the RL method yields a substantial reduction in RL2E for all cases, demonstrating significantly improved mean prediction accuracy relative to both the LF and DL results.
Among the RL cases, a slight increase is observed at 20\% noise, indicating that excessive noise perturbation begins to degrade deterministic accuracy.

The LPP results in \Cref{fig:lpp} provide further insight into the predictive capacity of the proposed framework.
Since LPP jointly evaluates prediction accuracy and predictive variance, an overconfident model with small uncertainty bands tends to produce a highly negative LPP value.
This behavior is clearly observed for the original RL result without added noise, which gives the most negative LPP among all cases, despite its strong mean accuracy. This confirms that the uncertainty estimate of the RL method is severely overconfident. As noise is introduced into the LF input, the LPP becomes markedly less negative, indicating improved probabilistic calibration. In particular, the 10\% and 15\% noise cases achieve the largest LPP values and are therefore the most favorable in terms of balancing predictive accuracy and uncertainty quantification.

The coverage results in \Cref{fig:coverage} show a consistent trend. The original RL result gives the smallest coverage, confirming that its uncertainty bands are too narrow and fail to enclose a sufficient portion of the true HF response. As the noise level increases, the empirical coverage increases monotonically, indicating that the uncertainty bands become wider and capture more of the true solution.
However, a larger coverage does not necessarily imply better overall performance, because overly wide uncertainty bands may trivially enclose the truth while sacrificing predictive accuracy.
Therefore, coverage should not be interpreted in isolation, but rather together with RL2E and LPP.

\Cref{fig:calibration_curve} presents the calibration curves for the DL and RL methods.
Each curve represents the relationship between the nominal coverage level and the corresponding empirical coverage computed from the predicted uncertainty intervals.
The black dashed diagonal line indicates perfect calibration.
Curves located below the diagonal indicate under-coverage, meaning that the predicted uncertainty intervals are too narrow and the model is overconfident.
In contrast, curves located above the diagonal indicate over-coverage, meaning that the predicted intervals are overly wide and the uncertainty estimates are conservative.
It can be observed that the RL model without noise is the most overconfident.
This indicates that the uncertainty estimated by the RL model without noise is insufficient to cover the true HF POD coefficients.
As the noise level increases, the calibration curve gradually moves closer to the ideal diagonal line, suggesting that noise injection improves the reliability of the quantified uncertainty.
Among the considered RL cases, the model with 15\% noise achieves the best calibration, with its curve being closest to the diagonal line over most nominal coverage levels.
When the noise level further increases to 20\%, the calibration curve moves above the diagonal line, suggesting that the uncertainty becomes slightly conservative due to over-expanded prediction intervals.
The calibration curve analysis demonstrates that an appropriate noise level can effectively improve uncertainty calibration, while excessive noise may lead to overly conservative uncertainty estimates.

Overall, the four metrics reveal a clear trade-off between deterministic accuracy and uncertainty calibration.
The DL method provides only limited improvement in global RL2E, although it improves some complex higher-order dynamics.
The RL method greatly improves the mean prediction accuracy, but without noise perturbation it is strongly overconfident.
Adding noise to the LF input effectively alleviates this issue by increasing predictive uncertainty and improving coverage, although excessive noise also reduces mean accuracy.
Considering all three metrics together, the RL method with 10\% or 15\% input noise appears to provide the best overall balance between accurate prediction and reliable uncertainty quantification.

\section{Conclusion}\label{sec:conclusion}

In this study, we proposed an uncertainty-aware MF closure correction framework based on CNFs.
The proposed framework investigates how CNF can be used for closure modeling while simultaneously providing uncertainty quantification.
This capability is important because closure modeling is inherently uncertain: unresolved dynamics, information loss in reduced-order representations, and error accumulation during recursive prediction can all lead to ambiguity in the low-to-high-fidelity mapping. 

To construct the MF framework, we employed Galerkin POD projection to define the LF ROM, and then trained a CNF to learn the discrepancy between the LF and HF representations in POD coefficients.
Two correction strategies were considered: In the DL formulation, the CNF directly learns the conditional mapping from the LF state to the corresponding HF state. In the RL formulation, the CNF instead learns the conditional distribution of the residual between the HF and LF states, which is then added back to the LF prediction to obtain the HF corrected state.

Two two-dimensional double shear layer and vortex merger problems were used to validate the proposed framework.
The results showed that the DL method provides some improvement in overall MF prediction accuracy, although it can still improve selected higher-order POD modes with complex dynamics.
By contrast, the RL method significantly improves mean prediction accuracy and more effectively captures the underlying reduced-order dynamics.
However, for the vortex merger problem, the uncertainty estimates produced by the original RL formulation were found to be overconfident.
To address this issue, Gaussian noise was introduced into the LF input, which helps propagate additional uncertainty through the correction process.
Numerical results demonstrate that this strategy alleviates overconfidence, with the 10\% and 15\% noise cases providing the most favorable balance between predictive accuracy and uncertainty calibration.

Although the results are promising, several important questions remain open for future study.
First, the role of injected noise in accounting for the uncertainty in the LF-to-HF map still requires deeper investigation.
Second, the optimal noise level is likely problem-dependent and may be influenced by the amount of training data, the complexity of the underlying dynamics, and the capacity of the CNF architecture.
In addition, future work should examine the robustness and generalizability of the framework for more complex flow problems, higher-dimensional systems, and different ROM settings.
Overall, this work demonstrates the potential of CNF-based probabilistic closure correction for ROM and provides a promising foundation for future developments in uncertainty-aware MF learning.

\section*{Acknowledgements}

This material is based upon work supported by the U.S. Department of Energy (DOE), Office of Science, Office of Advanced Scientific Computing Research, under the project "Uncertainty Quantification for Multifidelity Operator Learning (MOLUcQ)" (Project No. 81739).
The work of SEA was also supported by the U.S. DOE, Office of Science, Office of Advanced Scientific Computing Research through the Pacific Northwest National Laboratory Distinguished Computational Mathematics Fellowship (Project No. 71268) to provide guidance on fluid flow datasets and closure modeling framework.
Pacific Northwest National Laboratory is operated for DOE by Battelle Memorial Institute under Contract DE-AC05-76RL01830.

\bibliographystyle{elsarticle-num-names}
\bibliography{reference}

\end{document}